\documentclass[10pt,journal,compsoc]{IEEEtran}
\ifCLASSOPTIONcompsoc
  \usepackage[nocompress]{cite}
\else
  \usepackage{cite}
\fi

\ifCLASSINFOpdf
\else
\fi
\usepackage{amsmath,subcaption,multirow,amsfonts,amssymb}
\DeclareMathOperator*{\maxi}{maximize}
\DeclareMathOperator*{\mini}{minimize}
\DeclareMathOperator*{\argmax}{arg\,max}

\usepackage[ruled,linesnumbered]{algorithm2e}
\newcommand{\nonl}{\renewcommand{\nl}{\let\nl\oldnl}}
\makeatletter
\newcommand{\nosemic}{\renewcommand{\@endalgocfline}{\relax}}
\usepackage{graphicx}

\begin{document}


\title{Generative Adversarial Reward Learning for Generalized Behavior Tendency Inference}

%
%
%
%

\author{Xiaocong Chen,
        Lina Yao,~\IEEEmembership{Member,~IEEE,}
        Xianzhi Wang,~\IEEEmembership{Member,~IEEE,}
        Aixin Sun,~\IEEEmembership{Member,~IEEE,}
        Wenjie Zhang,~\IEEEmembership{Member,~IEEE,}
        Quan Z. Sheng,~\IEEEmembership{Member,~IEEE}
\IEEEcompsocitemizethanks{\IEEEcompsocthanksitem X. Chen, L. Yao and W. Zhang are with the School 
of Computer Science and Engineering, University of New South Wales, Sydney,
NSW, 2052, Australia.\protect\\
E-mail: \{xiaocong.chen, lina.yao, wenjie.zhang\}@unsw.edu.au
\IEEEcompsocthanksitem X. Wang is with School of Computer Science, University of Technology Sydney, Sydney, NSW, 2007, Australia.
\IEEEcompsocthanksitem A. Sun is with Nanyang Technological University, Singapore.
\IEEEcompsocthanksitem Q. Sheng is with Department of Computing, Macquarie University, Sydney, NSW, 2109, Australia.}
}

%
%

\markboth{Journal of \LaTeX\ Class Files,~Vol.~14, No.~8, August~2015}%
{Shell \MakeLowercase{\textit{et al.}}: Bare Demo of IEEEtran.cls for Computer Society Journals}
%



\IEEEtitleabstractindextext{%
\begin{abstract}
Recent advances in reinforcement learning have inspired increasing interest in learning user modeling adaptively through dynamic interactions, e.g.,
in reinforcement learning based recommender systems.
Reward function is crucial for most of reinforcement learning applications as it can provide the guideline about the optimization.
However, current reinforcement-learning-based methods rely on manually-defined reward functions, which cannot adapt to dynamic and noisy environments. Besides, they generally use task-specific reward functions that sacrifice generalization ability. We propose a generative inverse reinforcement learning for user behavioral preference modelling, to address the above issues. Instead of using predefined reward functions, our model can automatically learn the rewards from user's actions based on discriminative actor-critic network and Wasserstein GAN. Our model provides a general way of characterizing and explaining underlying behavioral tendencies, and our experiments show our method outperforms state-of-the-art methods in a variety of scenarios, namely traffic signal control, online recommender systems, and scanpath prediction. 
\end{abstract}

}

\maketitle

\IEEEdisplaynontitleabstractindextext

%
\IEEEpeerreviewmaketitle

\IEEEraisesectionheading{\section{Introduction}\label{sec:introduction}}
\IEEEPARstart{B}{ehavior} modeling provides a footprint about user's behaviors and preferences. It is a cornerstone of diverse downstream applications that support personalized services and predictive decision-making, such as human-robot interactions, recommender systems, and intelligent transportation systems.
Recommender systems generally use users' past activities to predict their future interest ~\cite{hong2009context,zheng2018drn,zhang2019deep}; past studies integrate demographic information with user's long-term interest on personalized tasks~\cite{hu2017diversifying,xu2020contextual,wang2014exploration,kim2003learning}.
In human-robot interaction, a robot learns from user behaviors to predict user's activities and provide necessary support~\cite{sheridan2016human}. Multimodal probabilistic models~\cite{schmerling2018multimodal} and teacher-student network~\cite{siam2019video} are often used to predict user's intention for traffic prediction or object segmentation.
Travel behavior analysis ; it is a typical task in smart-city applications~\cite{badii2017user,bellini2017wi}.

Traditional methods learn static behavioral tendencies via modeling user's historical activities with items as a feature space~\cite{seko2011group} or a user-item matrix~\cite{shi2014collaborative}.
In contrast, reinforcement learning shows advantages in learning user's preference or behavioral tendency through dynamic interactions between agent and environment. It has attracted lots of research interests in recommendation systems~\cite{wang2014exploration}, intention prediction~\cite{yang2020predicting}, traffic control~\cite{bazzan2009opportunities},  and human-robot interaction domains~\cite{liu2018interactive}.
Reinforcement learning covers several categories of methods, such as value-based methods, policy-based methods, and hybrid methods. All these methods use the accumulated reward during a long term to indicate user's activities. The reward function is manually defined and requires extensive effort to contemplate potential factors.  

In general, user's activities are noisy, occasionally contaminated by imperfect user behaviors and thus may not always reveal user's interest or intention.
For example, in online shopping, a user may follow a clear logic to buy items and randomly add additional items because of promotion or discounts.
This makes it difficult to define an accurate reward function because the noises also affect the fulfillment of task goals in reinforcement learning.
Another challenge lies in the common practice of adding task-specific terms to the reward function to cope with different tasks.
Current studies usually require manually adjusting the reward function to model user's profiles~\cite{zheng2018drn,chen2018stabilizing,chen2019large}. Manual adjustment tends to produce imperfect results because it is unrealistic to consider all reward function possibilities, not to mention designing reward functions for new tasks.

A better way to determine the reward function is to learn it automatically through dynamic agent-environment interactions. Inverse reinforcement learning recently emerges as an appealing solution, which learns reward function learning from demonstrations in a few scenarios~\cite{ng2000algorithms}.
However, inverse reinforcement learning faces two challenges for user behavior modeling. First, it requires a repeated, computational expensive reinforcement learning process to apply a learned reward function~\cite{ho2016generative}; second, given an expert policy, there could be countless reward functions for choice, making the selection of reward function difficult and the optimization computationally expensive. The only recommendation model~\cite{chen2020generative} that adopts improved inverse reinforcement learning simply skips the repeated reinforcement learning process; thus, it is hard to converge as it lacks sampling efficiency and training stability;
furthermore, the model only works for recommender systems and lacks generalization ability. 

With such existing challenges, manually designed reward function has less feasibility and generalizability. Moreover, \cite{chen2020generative} employs inverse reinforcement learning to learn the reward from demonstration which still suffers the undefined problem due to the nature of the logarithm. To relieve this, we manipulate the function by adding an extra learnable term to avoid such a problem. In addition, existing works do not consider the absorbing state problem such that agents will stop learning once the absorbing states are reached. The major reason why is that agent will receive zero rewards in absorbing states and may lead to a sub-optimal policy.

In this paper, we aim to construct user models directly from an array of various demonstrations efficiently and adaptively, based on a generalized inverse reinforcement learning method.
Learning from demonstrations not only avoids the need for inferring a reward function but also reduces computational complexity.
\begin{figure*}[!ht]
    \centering
    \includegraphics[width=\linewidth]{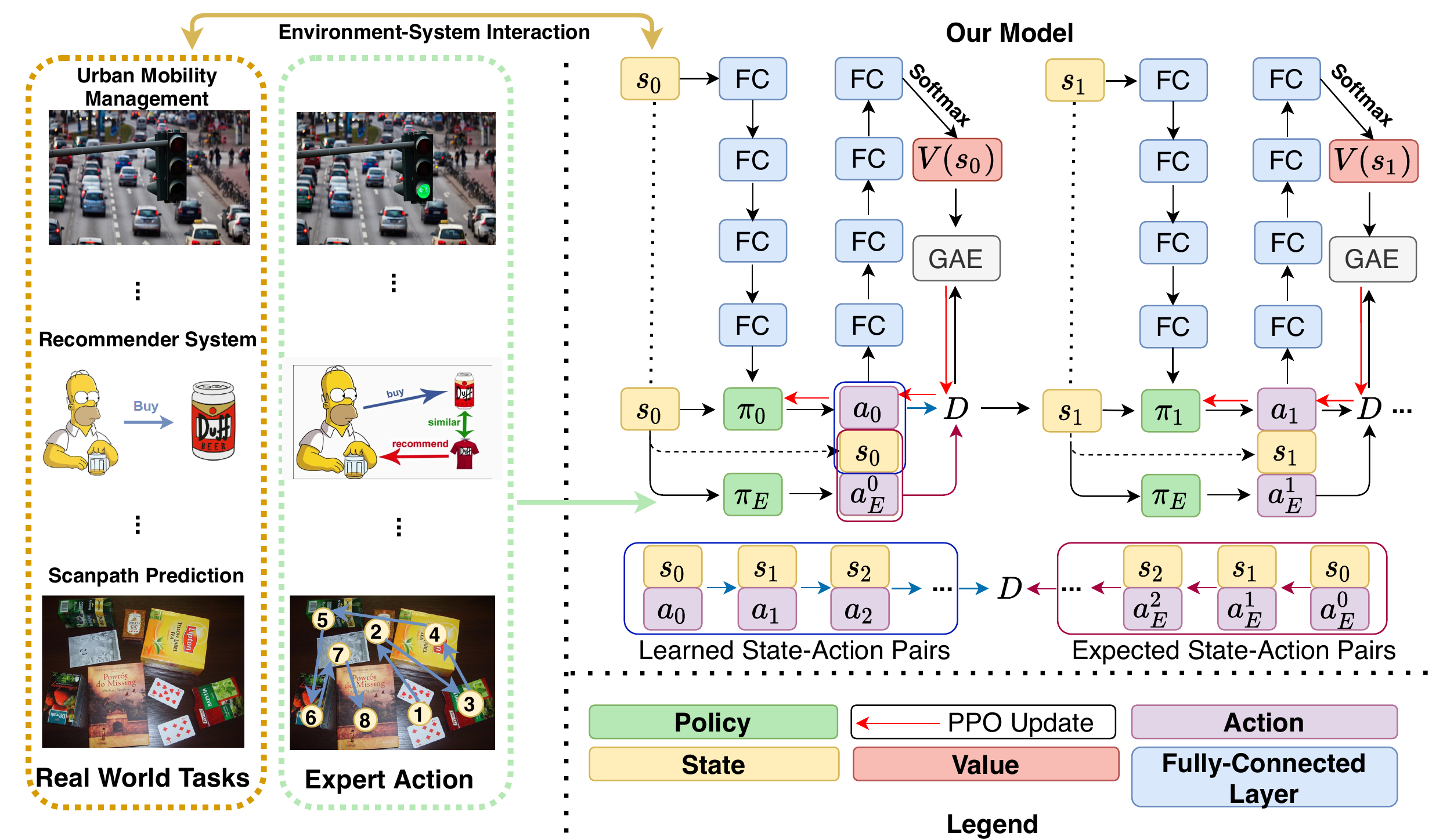}
    \caption{Overall structure of the proposed framework. The left-hand side provides three example environments from top to bottom: urban mobility management, recommender system and scanpath prediction. The proposed model will interact with the environment to achieve the corresponding state representations for current task. The expert actions will be achieved simultaneously and feed into our model to participate the training procedure of the discriminator. }
    \label{fig:str}
\end{figure*}
To this end, we propose a new model that employs a generative adversarial strategy to generate candidate reward functions and to approximate the true reward.
We use the new model as a general way of characterizing and explaining tendencies in user behaviors.
In summary, we make the following contributions:
\begin{itemize}
    \item We propose a new inverse reinforcement-learning-based method to capture user's behavioral tendencies. To the best of our knowledge, this is the first work to formulate user's behavioral tendency using inverse reinforcement learning.
    \item We design a novel stabilized sample-efficient discriminative actor-critic network with Wasserstein GAN to implement the proposed framework. Our framework is off-policy and can reduce interactions between system and environment to improve efficiency. Besides, we integrate a learnable term into our reward function to increase the capability of our method.
    \item Our extensive experiments demonstrate the generalization ability and feasibility of our approach in three different scenarios. We use visualization to show the explainability of our method.
\end{itemize}
\section{Problem Formulation and Preliminary}
Behavioral tendency refers to user's preferences at a certain timestamp and is usually hard to be evaluated directly.
The common way to evaluate behavioral tendencies is to examine how well the actions taken out of the learned behavioral tendencies match the real actions taken by the user.
It is similar to reinforcement learning's decision-making process, where the agent figures out an optimal policy $\pi$ such that each action of it could achieve a good reward.

In this work, we define behavioral tendencies modeling as an optimal policy finding problem.
Given a set of users $\mathcal{U} = \{u_0,u_1,\cdots,u_n\}$, a set of items $\mathcal{O}=\{o_0,o_1,\cdots,o_m\}$ and user's demographic information $\mathcal{D}=\{d_0,d_1,\cdots,d_n\}$. We first define the Markov Decision Process (MDP) as a tuple $(\mathcal{S},\mathcal{A},\mathcal{P},\mathcal{R},\gamma)$, where $\mathcal{S}$ is the state space (i.e., the combination of the subset of $\mathcal{O}$, subset of $\mathcal{U}$ and its corresponding $\mathcal{D}$).
$\mathcal{A}$ is the action space, which includes all possible agent's decisions,
$\mathcal{R}$ is a set of rewards received for each action $a\in\mathcal{A}$,
$\mathcal{P}$ is a set of state transition probability,
and $\gamma$ is the discount factor used to balance the future reward and the current reward.
The policy can be defined as $\pi:\mathcal{S}\rightarrow \mathcal{A}$---given a state $s\in\mathcal{S}$, $\pi$ will return an action $a\in \mathcal{A}$ so as to maximize the reward. However, it is unrealistic to find a universal reward function for user behavioral tendency, which is highly task-dependent. Hence, we employ
Inverse reinforcement learning (IRL) to learn a policy $\pi$ from the demonstration from expert policy $\pi_E$, which always results in user's true behavior. We formulate the IRL process using a uniform cost function $c(s,a)$~\cite{ng2000algorithms}: 
\begin{align}
    \mini_{\pi} \max_{c\in\mathcal{C}} \mathbb{E}_{\pi}[c(s,a)] - \mathbb{E}_{\pi_E}[c(s,a)]\label{eq1}
\end{align}
The cost function class $\mathcal{C}$ is restricted to convex sets defined by the linear combination of a few basis functions \{$f_1,f_2, \cdots, f_k$\}. Hence, given a state-action pair $(s,a)$, the corresponding feature vector can be represented as $f(s,a) = [f_1(s,a),f_2(s,a),\cdots,f_k(s,a)]$.
$\mathbb{E}_{\pi}[c(s,a)]$ is defined as (on $\gamma$-discounted infinite horizon):
\begin{align}
    \mathbb{E}_{\pi}[c(s,a)] = \mathbb{E}[\sum_{t=0}^{\infty}\gamma^t c(s_t,a_t)]
\end{align}



According to Eq.(\ref{eq1}), the cost function class $\mathcal{C}$ is convex sets, which have two different formats: linear format~\cite{abbeel2004apprenticeship} and convex format~\cite{syed2008apprenticeship}, respectively:
\begin{align}
    & \mathcal{C}_{l} = \Big\{\sum_i w_if_i: \|w\|_2 \leq 1\Big\} \\
    & \mathcal{C}_{c} = \Big\{\sum_i w_if_i: \sum_i w_i = 1, \forall i \text{ s.t. } w_i \geq 0\Big\}
\end{align}
The corresponding objective functions are as follows:
\begin{align}
    &  \|\mathbb{E}_\pi[f(s,a)] - \mathbb{E}_{\pi_E}[f(s,a)]\|_2  \label{eq2}\\ 
    &  \mathbb{E}_\pi[f_j(s,a)] - \mathbb{E}_{\pi_E}[f_j(s,a)] \label{eq3}
\end{align}

Eq.(\ref{eq2}) is known as feature expectation matching~\cite{abbeel2004apprenticeship}, which aims to minimize the $l_2$ distance between the state-action pairs that are generated by learned policy $\pi$ and expert policy $\pi_E$.
Eq.(\ref{eq3}) aims to minimize the function $f_j$ such that the worst-case should achieve a higher value~\cite{syed2008game}. Since Eq.(\ref{eq1} suffers the feature ambiguity problem, we introduce $\gamma$-discounted causal entropy~\cite{bloem2014infinite} (shown below) to relieve the problem:
\begin{align}
     H(\pi) \triangleq  & \mathbb{E}_{\pi}[-\log\pi(a|s)] =  \mathbb{E}_{s_t,a_t\sim\pi}\bigg[-\sum_{t=0}^\infty\gamma^t\log\pi(a_t|s_t)\bigg] \label{eq6}
\end{align}
As such, Eq.(\ref{eq1}) can be written by using the $\gamma$-discounted causal entropy as:
\begin{align}
    \mini_{\pi}-H(\pi) - \mathbb{E}_{\pi_E}[c(s,a)] + \max_{c\in\mathcal{C}}\mathbb{E}_{\pi}[c(s,a)] \label{eq7}
\end{align}

Suppose $\Pi$ is the policy set. We define the loss function $c(s,a)$ to ensure the expert policy receives the lowest cost while all the other learned policies get higher costs.
Referring to Eq.(\ref{eq7}), the maximum causal entropy inverse reinforcement learning~\cite{ziebart2010modeling} works as follows:
\begin{align}
    \maxi_{c\in\mathcal{C}} (\min_{\pi \in \Pi} -H(\pi) + \mathbb{E}_{\pi}[c(s,a)])  - \mathbb{E}_{\pi_E}[c(s,a)]  \label{eq9}
\end{align}

Then, the policy set $\Pi$ can be obtained via policy generation.
Policy generation is the problem of matching two occupancy measures and can be solved by training a Generative Adversarial Network (GAN)~\cite{goodfellow2014generative}.
The occupancy measure $\rho$ for policy $\pi$ can be defined as:
\begin{align}
    \rho_{\pi}(s,a) = \pi(s|a)\sum_{t=0}^\infty \gamma^t P(s_t=s|\pi) \label{eq10}
\end{align}
We adopts GAIL~\cite{ho2016generative} and make an analogy from the occupancy matching to distribution matching to bridge inverse reinforcement learning and GAN. A GA regularizer is designed to restrict the entropy function:
\begin{align}
    \psi_{GA}(c(s,a)) = 
    \begin{cases} 
      \mathbb{E}_{\pi_E}[-c(s,a)-\log(1-\exp(c(s,a)))] & c < 0 \\
      \infty & c\geq 0 \\
  \end{cases}
\end{align}

The GA regularizer enables us to measure the difference between the $\pi$ and $\pi_E$ directly without the reward function:
\begin{align}
    \psi_{GA}(\rho_\pi - \rho_{\pi_E}) = \max_{D\in(0,1)^{\mathcal{S}\times\mathcal{A}}}\mathbb{E}_{\pi}[\log D(s,a)] \nonumber \\ + \mathbb{E}_{\pi_E}[\log (1-D(s,a))] \label{eq11}
\end{align}

The loss function from the discriminator $D$ is defined as $c(s,a)$ in Eq.(\ref{eq9}); it uses negative log loss (commonly used for binary classification) to distinguish the policies $\pi$ and $\pi_E$ via state-action pairs. The optimal of Eq.(\ref{eq11}) is equivalence to the Jensen-Shannon divergence~\cite{nguyen2009surrogate}:
\begin{align}
    D_{JS}(\rho_\pi,\rho_{\pi_E}) = D_{KL}(\rho_\pi\|(\rho_\pi+\rho_{\pi_E})/2) \nonumber + \\ D_{KL}(\rho_{\pi_E}\|(\rho_\pi+\rho_{\pi_E})/2)  \label{eq12}
\end{align}

Finally, we rewrite inverse reinforcement learning by substituting the GA regularizer into Eq.(\ref{eq7}):
\begin{align}
    \mini_{\pi}-\lambda H(\pi) + \underbrace{\psi_{GA}(\rho_\pi - \rho_{\pi_E})}_{D_{JS}(\rho_\pi,\rho_{\pi_E})} \label{eq13}
\end{align}
where $\lambda$ is a factor with $\lambda \geq 0$.
Eq.(\ref{eq13}) has the same goal as the GAN, i.e., finding the squared metric between distributions.
Eq.(\ref{eq13}) can be further extended into the following, which serves as the objective function for GAIL:
\begin{align}
    & \mini_{\pi}-\lambda H(\pi) + \psi_{GA}(\rho_\pi - \rho_{\pi_E})  \equiv \nonumber \min_{\pi}\max_{D} \mathcal{L_D}\\
    & \mathcal{L_D}=\mathbb{E}_{\pi}[\log D(s,a)] + \mathbb{E}_{\pi_E}[\log (1-D(s,a))] -\lambda H(\pi) \label{eq14}
\end{align}
We summarized all the notations used in this paper in Table~\ref{tab:symbols}.
\begin{table}[t]
    \caption{Main notations}\smallskip
    \centering
    \begin{tabular}{c|l}
        \hline 
        Symbols & Meaning \\
        \hline 
         $\mathcal{U}$ & Set of users \\ 
         $\mathcal{O}$ & Set of items \\
         $\mathcal{R}$ & Set of rewards received \\
         $\mathcal{D}$ & Set of demographic information \\ 
         $|\cdot|$ & Number of unique elements in $\cdot$\\
         $\gamma $ & Discount Factor \\
         $H(\pi)$ & $\gamma$-discounted casual entropy \\
         $\mathbb{E}$ & Expectation  \\
         $\rho$ & Occupancy Measure \\
         $S_t$ & State space at timestamp $t$\\
         $a_t$ & Action space at timestamp $t$ \\
         $\pi$ & Policy \\
         $\pi_E$ & Expert Policy \\
         $D$ & Discriminator \\
         $D_{KL}$ & Kullback–Leibler divergence\\
         $D_{JS}$ & Jensen-Shannon divergence \\
         $\cdot \| \cdot $ & Divergence\\
         \hline
    \end{tabular}
    \label{tab:symbols}
\end{table}
\section{Methodology}
The overall structure of our proposed method (shown in Fig.~\ref{fig:str}) consists of three components: policy and reward learning, stabilized sample efficient discriminative actor-critic network, and its optimization. Policy and reward learning aims to solve the reward bias and the absorbing state problem by introducing a learnable reward function and environment feedback. The stabilized actor-critic network aims to improve the training stability and sample efficiency for the ex sting methods. Optimization refers to the method to optimize the policy and the algorithms to train the overall approach.

\subsection{Policy and Reward Learning}
We consider behavioral tendencies inference as an agent policy learning problem and an agent policy as the abstraction of user's behavioral tendencies.
Policy learning aims to make the learned policy $\pi$ and expert policy $\pi_E$.
We define the occupancy measure $\rho$   in Eq.(\ref{eq10}) and solve policy learning as a occupancy measure based distribution matching problem\cite{abbeel2004apprenticeship}.
To this end, we define a reward function below to determine the performance in existing methods:
\begin{align}
    r(s,a) = \log(D(s,a))-\log(1-D(s,a)) \label{eq17}
\end{align}
~\cite{fu2018learning} design a dynamic robust disentangled reward function for the approximation by introducing the future state $s'$.
\begin{align}
    r'(s,a) = \log(D(s,a,s'))-\log(1-D(s,a,s')) \label{eq18}
\end{align}

The reward function defined in Eq.(\ref{eq17}) is not robust for dynamic environments. Although Eq.(\ref{eq18}) improves it by assigning positive and negative rewards for each time step to empower the agent to fit into different scenarios, both Eq.(\ref{eq17}) and Eq.(\ref{eq18}) have the absorbing state problem, i.e., the agent will receive no reward at the end of each episode, leading to sub-optimal policies~\cite{kostrikov2018discriminatoractorcritic}.
Specifically, instead of exploring more policies, the reward function $r(s,a)$ will assign a negative reward bias for the discriminator to distinguish samples from the generated policies and expert policies at the beginning of the learning process. Since the agent aims to avoid the negative penalty, the zero reward may lead to early stops.

Moreover, the above two reward functions are more suitable for survival or exploration tasks rather than the goal of this study. For survival tasks, the reward used on GAIL is $\log D(s,a)$, which is always negative because $D(s,a)$ ($\in[0,1]$) encourages agent to end current episode to stop more negative rewards. For exploration tasks, the reward function is $-\log(1-D(s,a))$, it is always positive and may result in the agent to loop in the environment to collect more rewards. 

We add a bias term to the reward function $r(s,a)$, as defined by either Eq.(\ref{eq17}) or Eq.(\ref{eq18}) to overcome the reward bias. In addition, we introduce a new reward given by environment $r_e$ for reward shaping. Finally, we have the following:
\begin{align}
    r_n(s,a) = \lambda_i \Bigg(r(s,a) + \sum_{t=T+1}^\infty \gamma^{t-T}r(s_a,\cdot)\Bigg) \label{eq19} + r_e
\end{align}
where $r(s_a,\cdot)$ is a learnable reward function, which is trainable during the training process.
We also add a dimension to indicate whether the current state is an absorbing state or not (denoted by 1 or 0, respectively). Besides, we simply sample the reward from the replay buffer, considering the bias term is unstable in practice.

\subsection{Stabilized Sample Efficient Discriminative Actor-Critic Network}
The stabilized sample efficient discriminative actor-critic network aims to enable the agent to learn the policy efficiently.
We take a variant of the actor-critic network, \textit{advantage actor-critic network}~\cite{mnih2016asynchronous}, as the backbone of our approach.
In this network, the actor uses policy gradient and the critic's feedback to update the policy, and the critic uses Q-learning to evaluate the policy and provides feedback~\cite{konda2000actor}.

Given the state space at timestamp $t$, the environment determines a state $s_t$, which contains user's recent interest and demographic information embedded, via the actor-network~\cite{chen2020knowledge,liu2020end}.
The actor-network feeds the state $s_t$ to a network that has four fully-connected layers with ReLU as the activation function. The final layer of the network outputs a policy function $\pi$, which is parameterized by $\theta$. 
Then, the critic network takes two inputs: the trajectory $(s_t,a_t)$, and the current policy $\pi_{\theta_t}$ from the actor-network. We concatenate the state-action pair $(s_t,a_t)$ and feed it into a network with four fully-connected layers (with ReLU as the activation function) and a softmax layer. The output of the critic-network is a value $V(s_t,a_t)\in {\rm I\!R}$ to be used for optimization (to be introduced later). 

The discriminator $D$ is the key component of our approach. To build an end-to-end model that better approximates the expert policy $\pi_E$, we parameterize the policy with $\pi_\theta$ and clip the discriminator's output so that $D:\mathcal{S}\times\mathcal{A} \to (0,1)$ with weight $w$. The loss function of $D$ is denoted by $\mathcal{L}_D$.
Besides, we use Adam~\cite{kingma2014adam} to optimize weight $w$ (the optimization for $\theta$ will be introduced later).
We consider the discriminator $D$ as a local cost function provider to guide the policy update.
During the minimization of the loss function $\mathcal{L}_D$, i.e., finding a point $(\pi,D)$ for it, the policy will move toward expect-like regions (divided by $D$) in the latent space.

Like many other networks, Actor-critic network also suffers the sample inefficiency problem ~\cite{wang2016sample}, i.e., the agent has to conduct sampling from the expert policy distribution, given the significant number of agent-environment interactions needed to learn the expert policy during the training process.
In this regard, we use an off-policy reinforcement learning algorithm (instead of on-policy reinforcement learning algorithms) to reduce interactions with the environment. In particular, we introduce a replay buffer $\mathcal{R}$ to store previous state-action pairs; when training discriminator we sample the transition from the replay buffer $\mathcal{R}$ in off-policy learning (instead of sampling trajectories from a policy directly).
We thereby define the loss function as follows: 
\begin{align}
    \mathcal{L}_D=\mathbb{E}_{\mathcal{R}}[\log D(s,a)] + \mathbb{E}_{\pi_E}[\log (1-D(s,a))] -\lambda H(\pi) \label{eq20}
\end{align}

Eq.(\ref{eq20}) matches the occupancy measures between the expert and the distribution induced by $\mathcal{R}$. Instead of comparing the latest trained policy $\pi$ and expert policy $\pi_E$, it comprises a mixture of all policy distributions that appeared during training.
Considering off-policy learning has different expectation from on-policy learning, we use importance sampling on the replay buffer to balance it.
\begin{align}
    & \mathcal{L}_D=\mathbb{E}_{\mathcal{R}}\bigg[\frac{\rho_{\pi_\theta}(s,a)}{\rho_{\mathcal{R}}(s,a)}\log D(s,a)\bigg] \nonumber \\ 
    & + \mathbb{E}_{\pi_E}[\log (1-D(s,a))] -\lambda H(\pi) \label{eq21}
\end{align}

      
Considering GAN has the training instability problem~\cite{arjovsky2017wasserstein}, we employ the Wasserstein GAN~\cite{gulrajani2017improved} to improve the discriminator's performance. While a normal GAN minimizes JS-Divergence cannot measure the distance between two distributions, Wasserstein GANs uses the EM-distance and Kantorovich-Rubinstein duality to resolve the problem~\cite{villani2008optimal}.
\begin{align}
    & \mathbb{E}_{\pi}[\log D(s,a)] - \mathbb{E}_{\pi_E}[\log (D(s,a))] \nonumber \\
    & + \mathbb{E}_{\pi_E}[(\|\nabla D(s,a)\| - 1)^2] \label{eq22}
\end{align}

We further use gradient penalty to improve the training for Wasserstein GANs~\cite{gulrajani2017improved}, given the gradient penalty can improve training stability for JS-Divergence-based GANs \cite{lucic2018gans}. 
We thereby obtain the final loss function as follows:
\begin{align}
    \mathcal{L}_D=\mathbb{E}_{\mathcal{R}}\bigg[\frac{\rho_{\pi_\theta}(s,a)}{\rho_{\mathcal{R}}(s,a)}\log D(s,a)\bigg] + \mathbb{E}_{\pi_E}[\log (1-D(s,a))] \nonumber \\ -\lambda H(\pi) + \mathbb{E}_{\pi_E}[(\|\nabla D(s,a)\| - 1)^2] \label{eq23}
\end{align}

\subsection{Optimization}
We conduct a joint training process on the policy network (i.e., the actor-critic network) and the discriminator. We parameterize the policy network with policy parameter $\theta$ and update it using trust region policy optimization (TRPO)~\cite{schulman2015trust} based on the discriminator.
TRPO introduces a trust region by restricting the agent's step size to ensure a new policy is better than the old one. We formulate the TRPO problem as follows:
\begin{align}
    \max_{\theta}\frac{1}{T}\sum_{t=0}^{T}\bigg[\frac{\pi_{\theta}(a_t|s_t)}{\pi_{\theta_{old}}(a_t|s_t)} A_t\bigg] \nonumber \\
    \text{subject to } D_{KL}^{\theta_{old}}(\pi_{\theta_{old}},\pi_{\theta}) \leq \eta  \label{eq24}
\end{align}
where $A_n$ is the advantage function calculated by Generalized Advantage Estimation (GAE) ~\cite{schulman2015high}. GAE is described as follows:
\begin{align}
     & A_t = \sum_{l=0}^\infty (\gamma \lambda_g)^l \delta_{t+l}^V \nonumber \\
     & \text{  where  } \delta_{t+l}^V = -V(s_t) + \sum_{l=0}^\infty \gamma^l r_{t+l} \label{eq25}
\end{align}
where $r_{t+l}$ is the test reward for $l$-step's at timestamp $t$, as defined on Eq.(\ref{eq19}).
Considering the high computation load of updating TRPO via optimizing Eq.(\ref{eq24}),
we update the policy using a simpler optimization method called Proximal Policy Optimization (PPO)~\cite{schulman2017proximal}, which has an objective function below:
\begin{align}
    & \underset{\tau\sim\pi_{old}}{\mathbb{E}}\Big[\sum_{t=0}^T\min\Big(\frac{\pi_{\theta}(a_t|s_t)}{\pi_{\theta_{old}}(a_t|s_t)}A_t,\nonumber\\
    & \text{clip}\Big(\frac{\pi_{\theta}(a_t|s_t)}{\pi_{\theta_{old}}(a_t|s_t)},1-\epsilon,1+\epsilon\Big)A_t\Big)\Big] \label{eq26}
\end{align}
where $\epsilon$ is the clipping parameter representing the maximum percentage of change that can be made by each update.

The overall training procedure is illustrated in Algorithm ~\ref{alg:d}, which involves the training of both the discriminator and the actor-critic network.
For the discriminator, we use Adma as the optimizer to find the gradient for Eq.(\ref{eq23}) for weight $w$ at step $i$:
\begin{align}
    \mathbb{E}_{\pi}[\nabla_w\log(D_w(s,a))] + \mathbb{E}_{\pi_E}[\nabla_w\log(1-D_w(s,a))] \nonumber \\
    + \mathbb{E}_{\pi_E}[(\|\nabla_w D(s,a)\| - 1)^2 \label{eq27}
\end{align}

\begin{algorithm}[ht]
\SetAlgoLined
\SetKwProg{Fn}{Function}{ is}{end}
 \SetKwInOut{Input}{input}
 \Input{Expert replay buffer $\mathcal{R}_E$, Initialize Policy Replay Buffer $\mathcal{R}$,Initialize policy parameter $\theta_0$, clipping parameter $\epsilon$}
 \Fn{Absorbing($\tau)$}{
    \If{$s_T$ is a absorbing state}{
        $\{s_T,a_T,\cdot, s'_T\} \leftarrow \{s_T,a_T,\cdot, s_a\} $\;
        $\tau\leftarrow\tau \cup \{s_a,\cdot,\cdot,s_a\}$\;
    }
    return $\tau$\;
    }
 \For{$\tau = \{s_t,a_t,\cdot,s'_t\}_{t=1}^T \in \mathcal{R}_E$}{
    $\tau\leftarrow$ Absorbing($\tau$)\;
 }
 $\mathcal{R}\leftarrow \emptyset$ \;
 \For{$i= 1,2,\cdots$}{
  Sampling  trajectories $\tau =\{s_t,a_t,\cdot,s'_t\}_{t=1}^T \sim\pi_{\theta_i}$\; 
  $\mathcal{R} \leftarrow \mathcal{R} \cup$ Absorbing($\tau$) \; 
  \For{$j = 1,\cdots,|\tau|$}{
      $\{s_t,a_t,\cdot,\cdot\}_{t=1}^B \sim \mathcal{R}$, $\{s'_t,a'_t,\cdot,\cdot\}_{t=1}^B \sim \mathcal{R}_E$ \;
      Update the parameter $w_i$ by gradient on Eq.(\ref{eq27}) \; 
  }
  \For{$j = 1,\cdots,|\tau|$}{
    $\{s_t,a_t,\cdot,\cdot\}_{t=1}^B \sim \mathcal{R}$\;
        \For{$b= 1, \cdots, B$}{
            $r=\log(D_{w_i}(s_b,a_b)) - \log(1 - D_{w_i}(s_b,a_b))$ \;
             Calculate the reshape reward $r'$ by Eq.(\ref{eq19}) \; 
            $(s_b,a_b,\cdot,s'_b) \leftarrow (s_b,a_b,r',s'_b)$ \;
        }
        \For{$k= 0,1,\cdots$}{ 
            Get the trajectories $(s,a)$ on policy $\pi_\theta = \pi(\theta_k)$ \;
            Estimate advantage $A_t$ using Eq.(\ref{eq25})\;
            \nosemic Compute the Policy Update \;
            \nonl\begin{equation*}
            \theta_{k+1} = \argmax_{\theta} \text{Eq.}(\ref{eq26})\;
            \end{equation*}
            \nonl By taking $K$ step of minibatch SGD (via Adma)\;
        }
        $\theta_i \leftarrow \theta_K$\;
  }
 }
 \caption{Training algorithm for our model}
 \label{alg:d}
\end{algorithm}

\begin{algorithm}[ht]
\SetAlgoLined
 \SetKwInOut{Input}{input}
 \Input{Initialize policy parameter $\theta_0$, clipping parameter $\epsilon$}
  \For{$k= 0,1,\cdots$}{ 
    Get the trajectories $(s,a)$ on policy $\pi_\theta = \pi(\theta_k)$ \;
    Estimate advantage $A_t$ using Eq.(\ref{eq25})\;
    \nosemic Compute the Policy Update \;
    \nonl\begin{equation*}
    \theta_{k+1} = \argmax_{\theta} \text{Eq.}(\ref{eq26})\;
    \end{equation*}
    \nonl By taking $K$ step of minibatch SGD (via Adma)\;
  }
    $\theta_i \leftarrow \theta_K$\;

 \caption{PPO Update}
 \label{alg:ppo}
\end{algorithm}

\section{Experiments}
We evaluate the proposed framework and demonstrate its generalization capability by conducting experiments in three different environments: Traffic Control, Recommendation System, and Scanpath Prediction.
Our model is implemented in Pytorch~\cite{paszke2019pytorch} and all experiments are conducted on a server with 6 NVIDIA TITAN X Pascal GPUs, 2 NVIDIA TITAN RTX with 768 GB memory.

\subsection{Urban Mobility Management}
In the traffic control scenario, the agent is required to control cars to conduct a certain task. The objective is to minimize the total waiting time in the trip. 
\begin{figure*}[htb]
    \begin{minipage}[t]{0.32\linewidth}
        \centering
        \includegraphics[width=\linewidth]{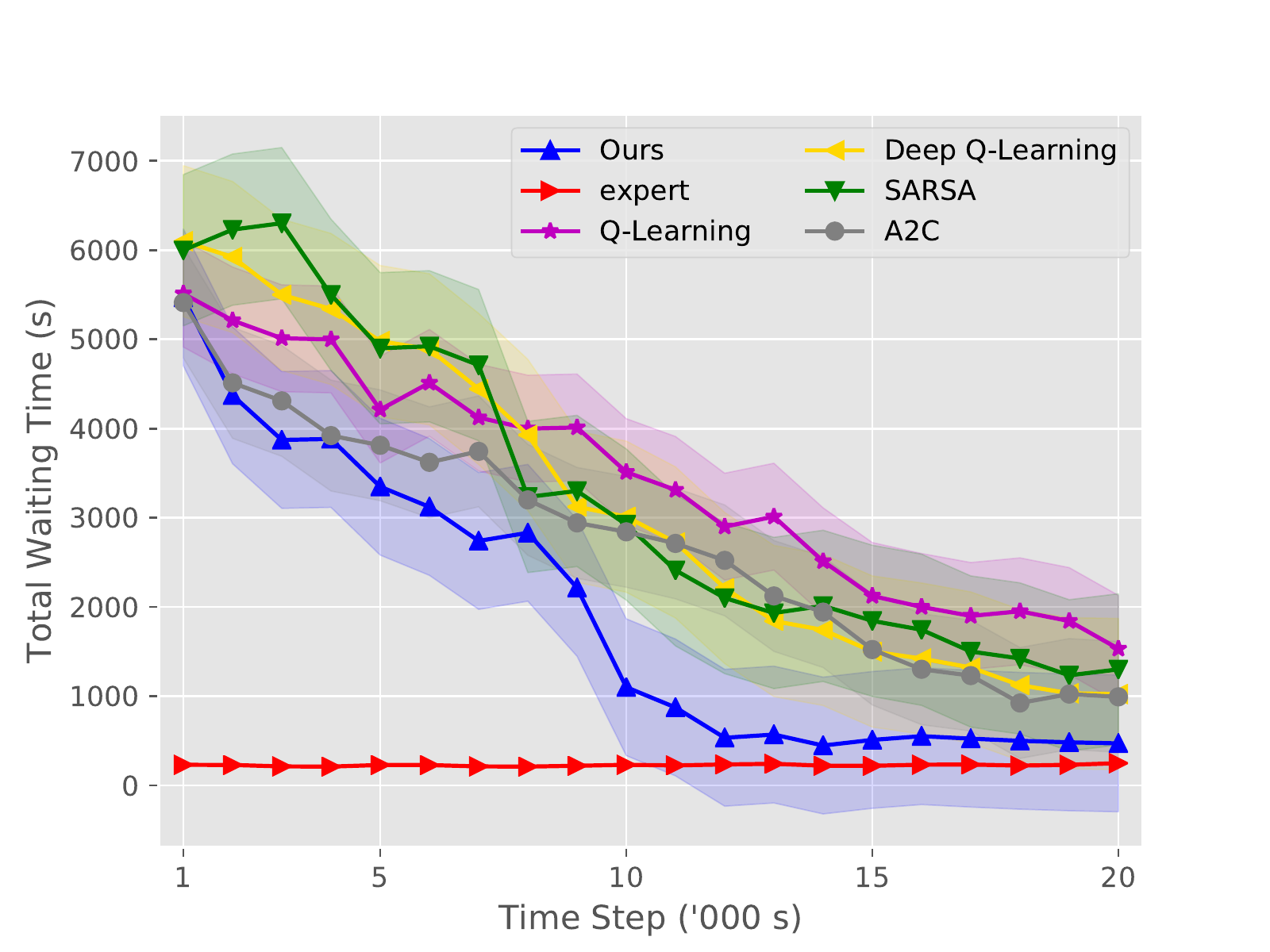}
        \subcaption{Total waiting time with 95\% confidence interval}
    \end{minipage}
    \hfill
    \begin{minipage}[t]{0.32\linewidth}
        \centering
        \includegraphics[width=\linewidth]{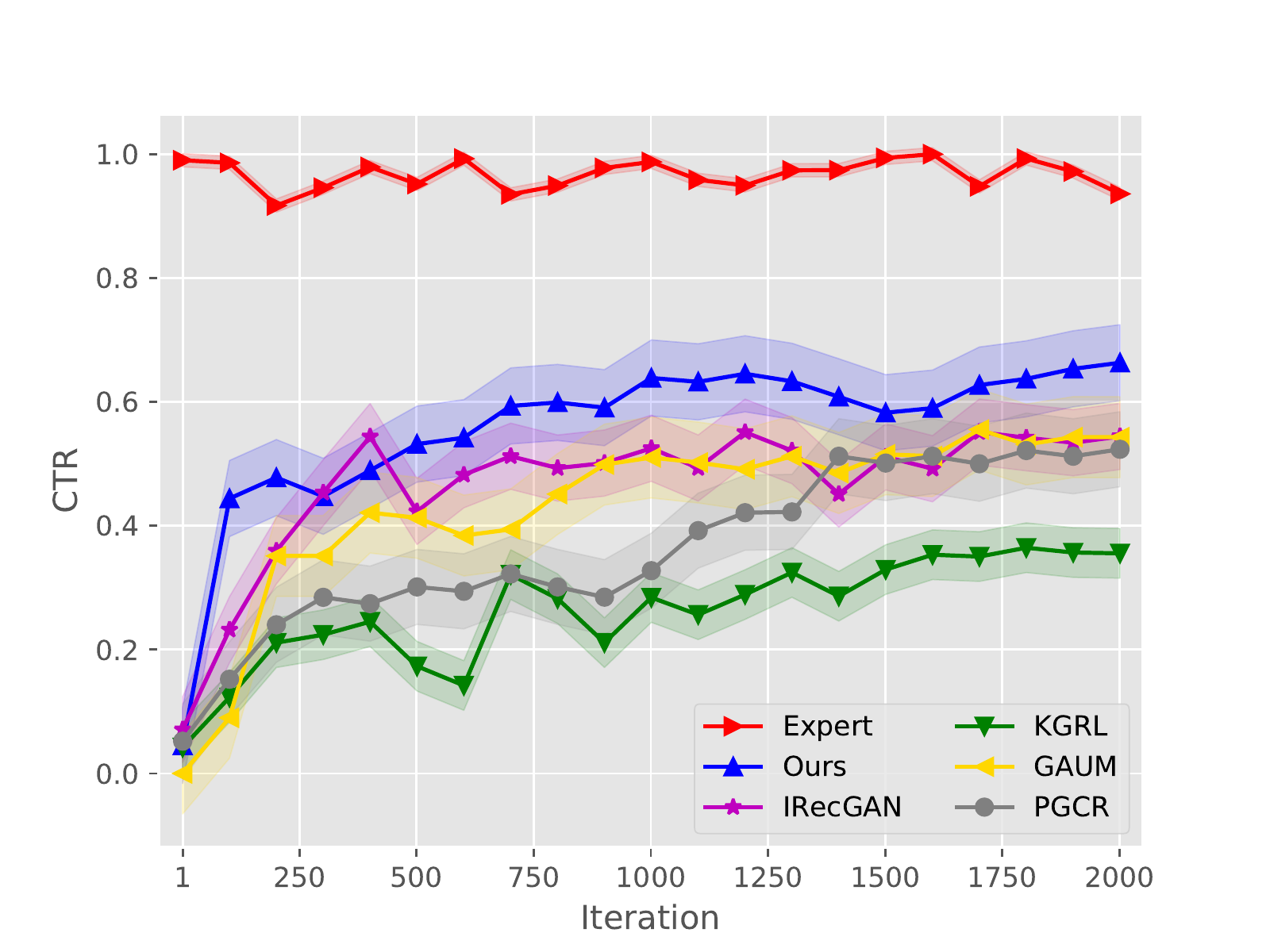}
        \subcaption{CTR with 95\% confidence interval}
    \end{minipage}
    \hfill
    \begin{minipage}[t]{0.32\linewidth}
        \centering
        \includegraphics[width=\linewidth]{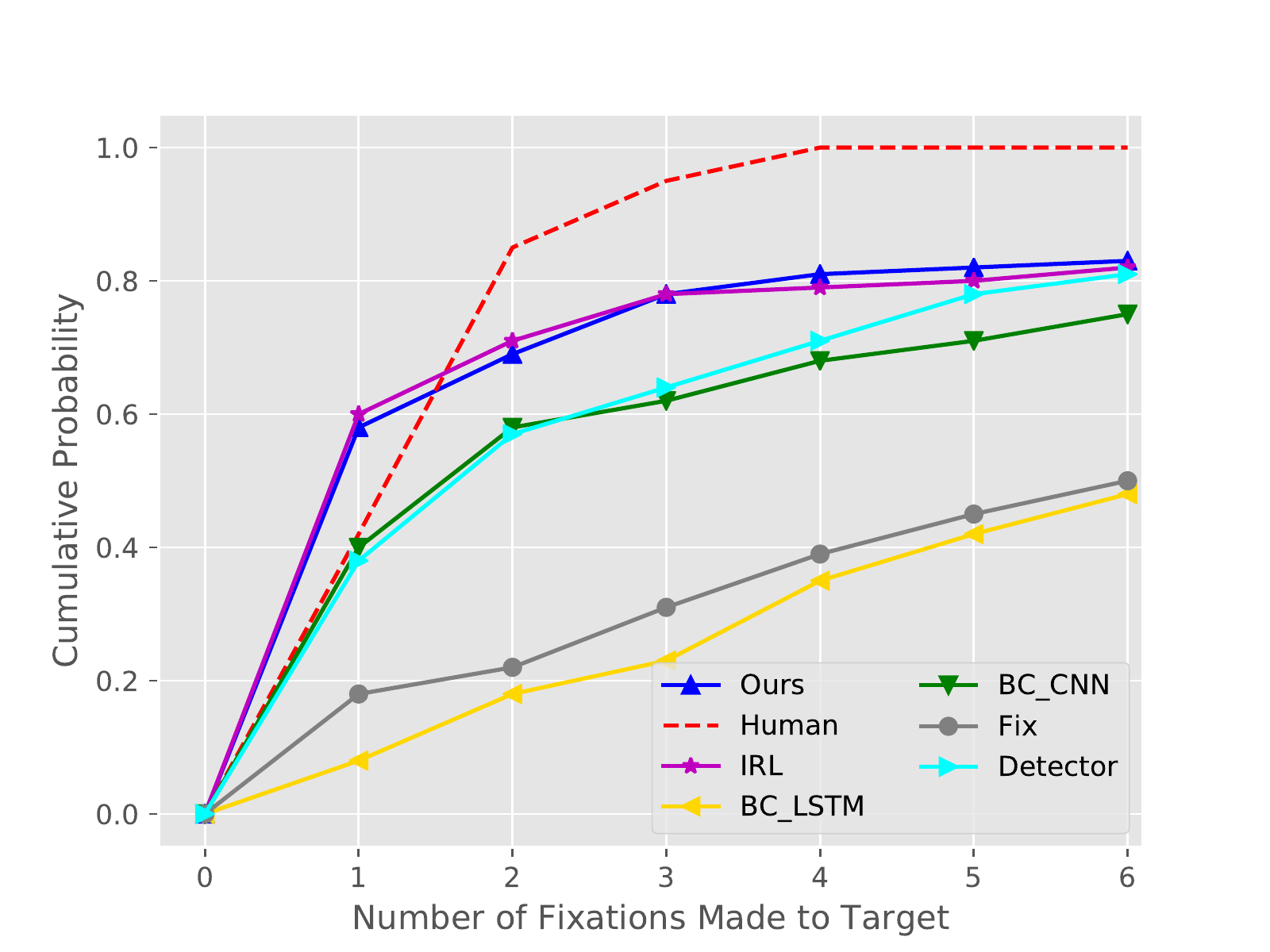}
        \subcaption{Cumulative probability comparison for selected baseline methods}
    \end{minipage}
    \caption{Overall results comparison. From left to right are represented to (a) Traffic Control, (b) Recommendation System and (c) Scanpath Prediction.}
    \label{fig:result}
\end{figure*}
\\
\subsubsection{Simulation of Urban Mobility} ~\\
Traffic signal control is critical to effective mobility management in modern cities. To apply our model to this context, we use the Simulation of Urban MObility (SUMO)~\cite{SUMO2018} library, a microscopic, space-continuous, and time-discrete traffic flow simulation tool, to test the method's performance. The agent controls traffic signals, and a car may take three actions facing traffic lights: go straight, turn left, or turn right, depending on user's preference.
We design a simple two-way road network that contains eight traffic lights for testing.
We employ an open-sourced library sumo-rl~\footnote{https://github.com/LucasAlegre/sumo-rl} to enable our agent can interact with the simulation environment (including receiving the reward) directly. The number of cars available in the environment is unlimited; the environment keeps generating cars until this simulation step ends or the road reaches its full capacity.\\

\subsubsection{Expert Policy Acquisition} ~\\ 
Since there is no official expert policy available for our customized road network, we use the same strategy as introduced by ~\cite{gao2018reinforcement} to collect a set of imperfect expert policies from a pre-trained policy network. This policy network is built upon actor-critic network, which is trained by using Deep Deterministic Policy Gradients (DDPG)~\cite{lillicrap2015continuous}. Expert policies are stored via state-action pairs, which concatenate observed states and expert actions.\\

\subsubsection{Baseline Methods} ~\\ 
We evaluate our model against several traditional reinforcement learning methods in this scenario. 
\begin{itemize}
    \item Q-Learning: An off-policy reinforcement learning method that finds the best action given the current state.
    \item Deep Q-learning: A deep Q-learning method that employs the neural network to extract features.
    \item SARSA~\cite{rummery1994line}: State–action–reward–state–action (SARSA) is an improved Q-learning method commonly used for traffic signal control.
    \item Advantage Actor-Critic Network (A2C)~\cite{mnih2016asynchronous}: An asynchronous method built on an actor-critic network for deep reinforcement learning.
\end{itemize}
Experiments are conducted in exactly the same environment to ensure a fair comparison. All the baseline methods are implemented by using PyTorch and are publicly available \footnote{https://github.com/hill-a/stable-baselines}. The reward provided by environment for each simulation step can defined as:
\begin{align}
    r = \sum_{n=0}^{N_{ts}} s * N_{cp}
\end{align}
where $N_{ts}$ is the number of traffic signals available in the environment, $s$ is the average car speed in this simulation step, and $N_{cp}$ is number of cars passed this traffic signals at the end of this simulation step. The evaluate metric is the total waiting time defined below:
\begin{align}
    t = \sum_{i=0}^{1000}\sum_{c=0}^{N_{c}} t_{c}^{i}
\end{align}
where $t_c^i$ is the time that the car $c$ waits at traffic light $i$, and $1,000$ is the duration for one simulation step. If car $c$ does not meet traffic light $i$, we set $t_{c}^{i} = 0$.
\\
\subsubsection{Hyper-parameters Setting and Results}  ~\\
DDPG parameters for the pre-trained model include $\gamma = 0.95, \tau=0.001$, the size of the hidden layer $128$, the size of the reply buffer $1,000$, and the number of episode $20,000$. Parameters for Ornstein-Uhlenbeck Noise include the scale $0.1$, $\mu=0, \theta=0.15, \sigma=0.2$.
For our method, we set the number of time steps to $20,000$, the hidden size of the advantage actor-critic network to $256$, the hidden size for discriminator to $128$, the learning rate to $0.003$, factor $\lambda$ to $10^{-3}$, mini batch size to $5$, and the epoch of PPO to $4$.
For the generalized advantage estimation, we set the discount factor $\gamma$ to $0.995$, $\lambda_g = 0.97$, and $\epsilon = 0.2$.
We also set $\lambda_i = 1$ for reward shaping and $\lambda = 1$ for $H(\pi)$.
The results in Fig.~\ref{fig:result} (a) show our method generally outperforms than all baseline methods.

\subsection{Recommendation System}
In the recommendation scenario, the agent aims to interact with a dynamic environment to mine user's interests and make recommendations to users.

\subsubsection{VirtualTB} ~\\ 
We use an open-source online recommendation platform, VirtualTB~\cite{shi2019virtual}, to test the performance of the proposed methods in a recommendation system.
VirtualTB is a dynamic environment built on OpenAI Gym\footnote{https://gym.openai.com/} to test our method's feasibility on recommendation tasks.
VirtualTB employs a customized agent to interact with it and achieves the corresponding rewards. It can also generate several customers with different preferences during the agent-environment interaction.
In VirtualTB, each customer has 11 static attributes encoded into an 88-dimensional space with binary values as the demographic information.
The customers have multiple dynamic interests, which are encoded into a 3-dimensional space and may change over the interaction process.
Each item has several attributes (e.g., price and sales volume), which are encoded into a 27-dimensional space. We use CTR as the evaluation metric because the gym environment can only provide rewards as feedback. CTR is defined as follows:
\begin{align}
    CTR = \frac{r_{episode}}{10*N_{step}}
\end{align}
where $r_{episode}$ is the reward that the agent receives at each episode. At each episode, the agent may take $N_{step}$ steps and receive a maximum reward of $10$ per step.
\\
\subsubsection{Baseline Methods} ~\\ 
We evaluate our model against four state-of-the-art methods covering methods based on deep Q-learning, policy gradient, and actor-critic networks.
\begin{itemize}
    \item IRecGAN~\cite{bai2019model}: An online recommendation method that employs reinforcement learning and GAN.
    \item PGCR~\cite{pan2019policy}: A policy-Gradient-based method for contextual recommendation.
    \item GAUM~\cite{chen2019generative}: A deep Q-learning based method that employs GAN and cascade Q-learning for recommendation.
    \item KGRL~\cite{chen2020knowledge}: An Actor-Critic-based method for interactive recommendation, a variant of online recommendation.
\end{itemize}
Note that GAUM and PGCR are not designed for online recommendation, and KGRL requires a knowledge graph---which is unavailable to the gym environment---as the side information, . Hence, we only keep the network structure of those networks when testing them on the VirtualTB platform.\\

\subsubsection{Hyper-parameters Setting and Results}~\\ 
The hyper-parameters are set in a similar way as in the traffic signals control. We set the number of episodes to $200,000$ for both the pre-trained policy network and our method. To ease comparison, we configure each iteration to contain 100 episodes.
The results in Fig.~\ref{fig:result} (b) show our method outperform all state-of-the-art methods. KGRL's poor performance may be partially caused by its reliance on knowledge graph, which is unavailable in our experiments.
\subsection{Scanpath Prediction}
Scanpath prediction is a type of goal-directed human intention prediction problem~\cite{yang2020predicting}. Take the last task in Fig.~\ref{fig:str} for example. Given a few objects, a user may first look at item 1, then follows the item numbers annotated in the figure, and finally reaches item 8. The task aims to predict user's intention (i.e., item 8), given the start item (i.e., item 1).\\ 
\subsubsection{Experimental Setup}~\\ 
We follow the same experimental setup as~\cite{yang2020predicting} and conduct all experiments on a public COCO-18 Search dataset\footnote{https://saliency.tuebingen.ai/datasets/COCO-Search18/index\_new.html}.
We replace the fully-connected layer in actor-network with CNN to achieve the best performance of our method on images. The critic-network has a new structure with two CNN layers followed by two fully-connected layers. The discriminator contains all CNN layers with a softmax layer as output. We also resize the input image from the original size of $1680\times1050$ into $320 \times 512$ and construct the state by using the contextual beliefs calculated from a Panoptic-FPN with a backbone network (ResNet-50-FPN) pretrained on COCO2017.
\\
\begin{figure*}[ht]
    \centering
    \begin{subfigure}{0.33\linewidth}
        \includegraphics[width=\linewidth]{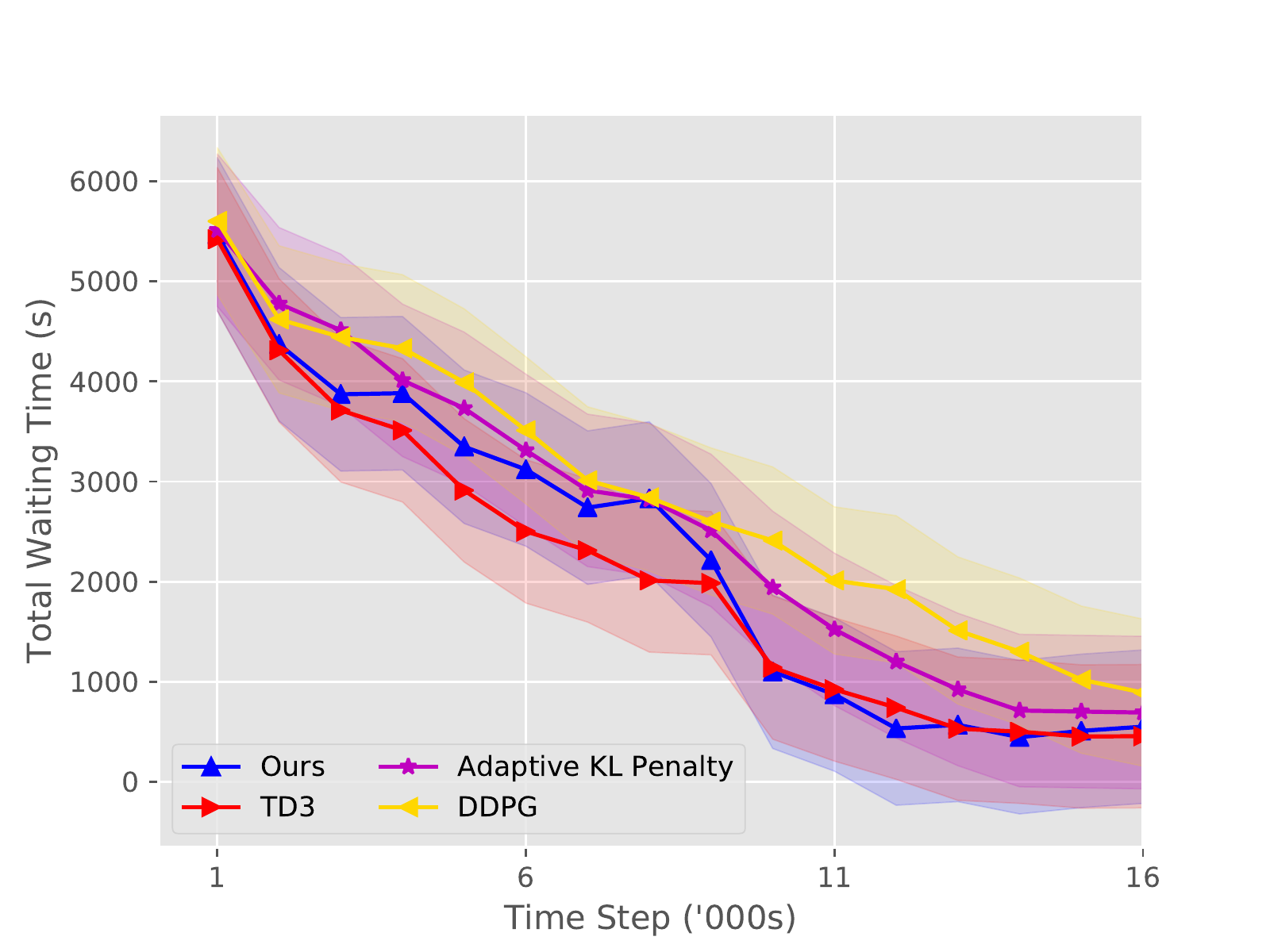}
        \caption{Traffic}
    \end{subfigure}
    \begin{subfigure}{0.33\linewidth}
        \includegraphics[width=\linewidth]{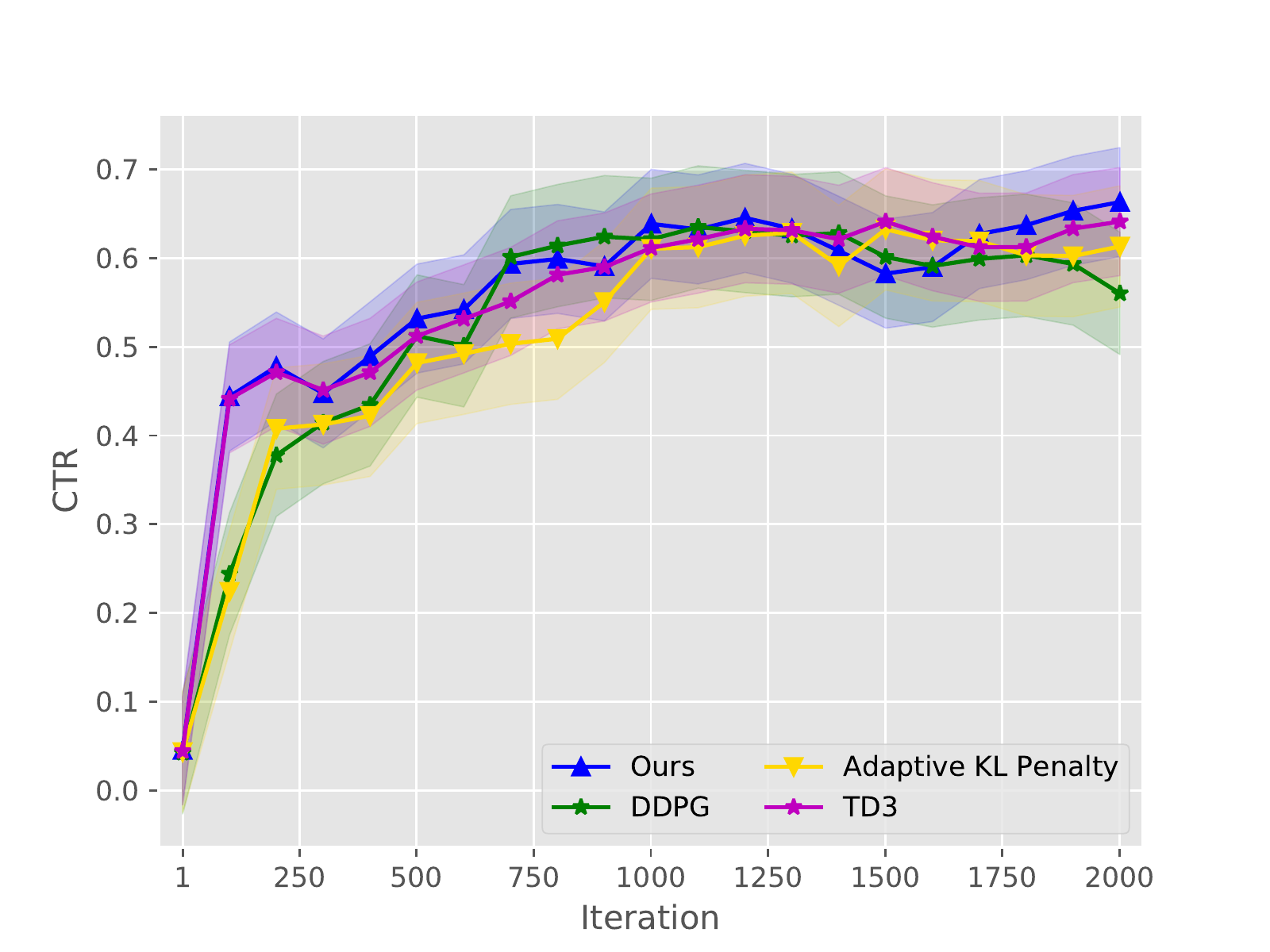}
        \caption{Recommendation}
    \end{subfigure}
    \begin{subfigure}{0.33\linewidth}
        \includegraphics[width=\linewidth]{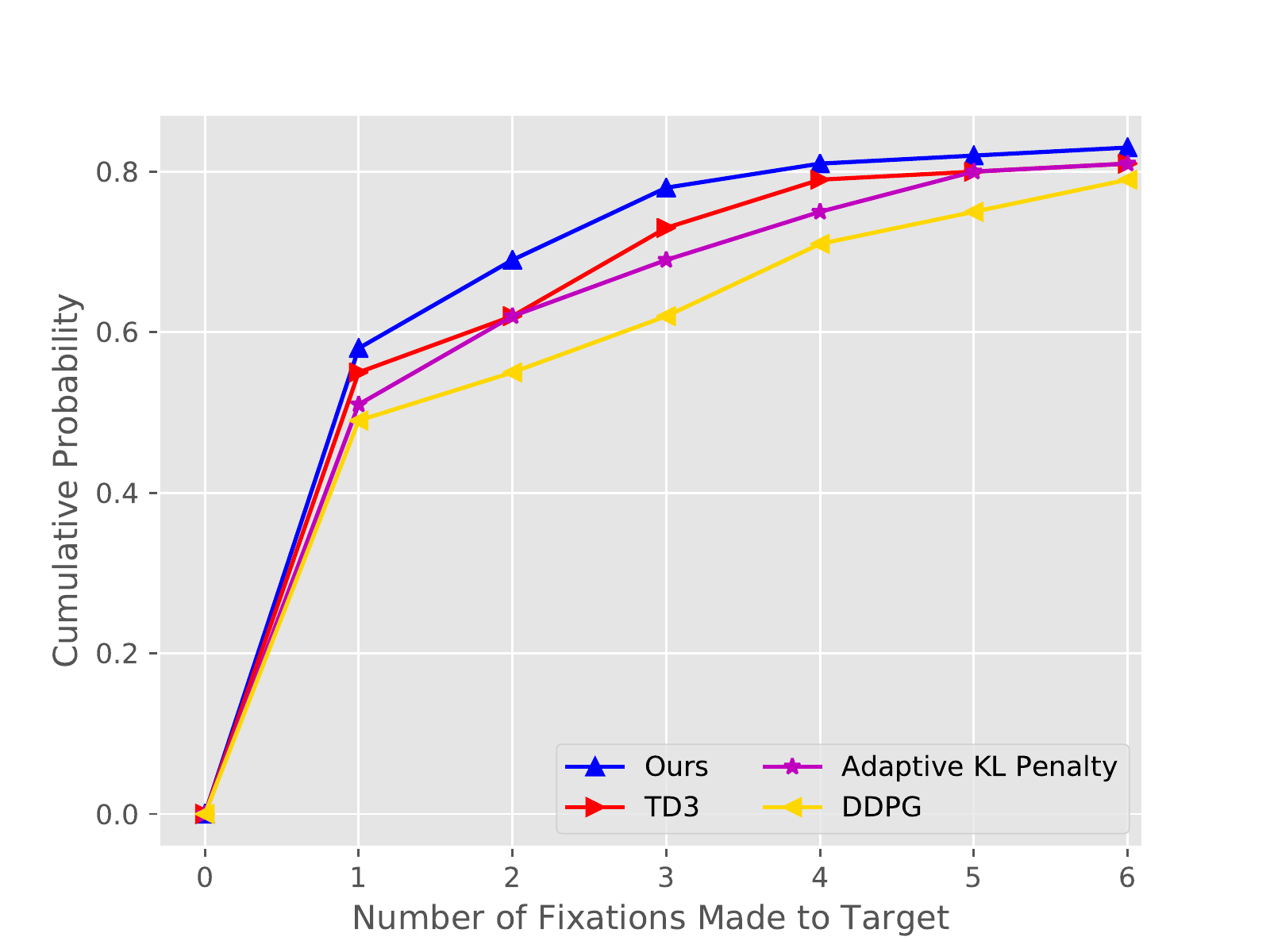}
        \caption{Scanpath Prediction}
    \end{subfigure}
    \caption{Results of ablation study for those three selected environments.}
    \label{fig:compare}
\end{figure*}
\subsubsection{Baseline Methods}~\\ 
We compare our method with several baseline methods, including simple CNN based methods, behavior-cloning based methods, and inverse reinforcement-learning-based method.
\begin{itemize}
    \item Detector: A simple CNN to predict the location of a target item.
    \item Fixation heuristics~\cite{yang2020predicting}: A method similar to \textit{Detector} but using the fixation to predict the location of a target item.
    \item BC-CNN~\cite{chen2015deepdriving}: A behavior-cloning method that uses CNN as the basic layer structure.
    \item BC-LSTM~\cite{ballas2015delving}: A behavior-cloning method that uses LSTM as the basic layer structure.
    \item IRL~\cite{yang2020predicting}: A state-of-the-art inverse reinforcement-learning-based method for scanpath prediction.
\end{itemize}
Experiments are conducted under the same conditions to ensure fairness. 
\\

\subsubsection{Performance Comparison}~\\
The hyper-parameters settings are the same as those used for the recommendation task.
We also use the same evaluation metrics as used in~\cite{yang2020predicting} to evaluate the performance: cumulative probability, probability mismatch, and scanpath ratio. The results in Fig.~\ref{fig:result}(c) show the cumulative probability of the gaze landing on the target after first six fixations. We report the probability mismatch and scanpath ratio in table~\ref{tab:path}.
\begin{table}[h]
    \centering
    \caption{Results comparison for selected methods on probability mismatch and scanpath ratio}
    \begin{tabular}{c|c|c}
        \hline
         &  Probability Mismatch $\downarrow$ & Scanpath Ratio$\uparrow$\\ \hline
         Human & n.a. & 0.862\\ \hline 
         Detector & 1.166 & 0.687\\
         BC-CNN &  1.328 & 0.706\\
         BC-LSTM & 3.497 & 0.406 \\
         Fixation & 3.046 & 0.545\\
         IRL & 0.987 & 0.862 \\ \hline
         Ours & \textbf{0.961} & \textbf{0.881} \\ \hline
    \end{tabular}
    \label{tab:path}
\end{table}

\subsection{Evaluation on Explainability}
\begin{figure}[h]
    \centering
    \begin{subfigure}{0.32\linewidth}
        \includegraphics[width=\linewidth]{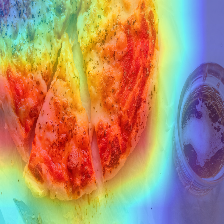}
    \end{subfigure}
    \begin{subfigure}{0.32\linewidth}
        \includegraphics[width=\linewidth]{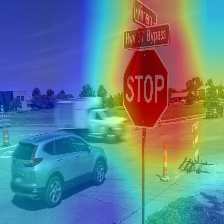}
    \end{subfigure}
        \begin{subfigure}{0.32\linewidth}
        \includegraphics[width=\linewidth]{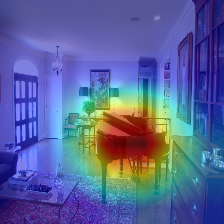}
    \end{subfigure}
    
    \begin{subfigure}{0.32\linewidth}
        \includegraphics[width=\linewidth]{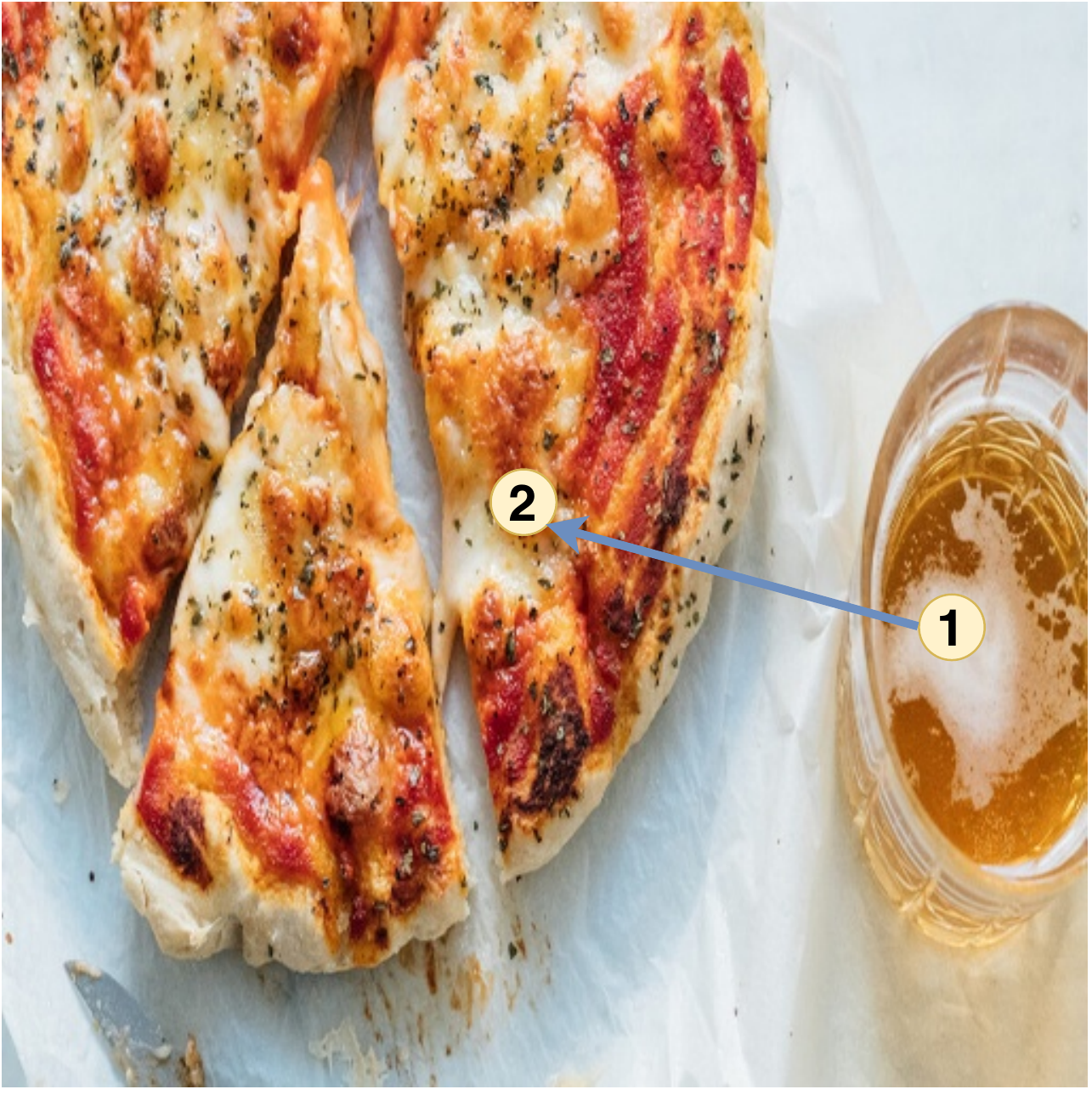}
        \caption{Piazza}
    \end{subfigure}
    \begin{subfigure}{0.32\linewidth}
        \includegraphics[width=\linewidth]{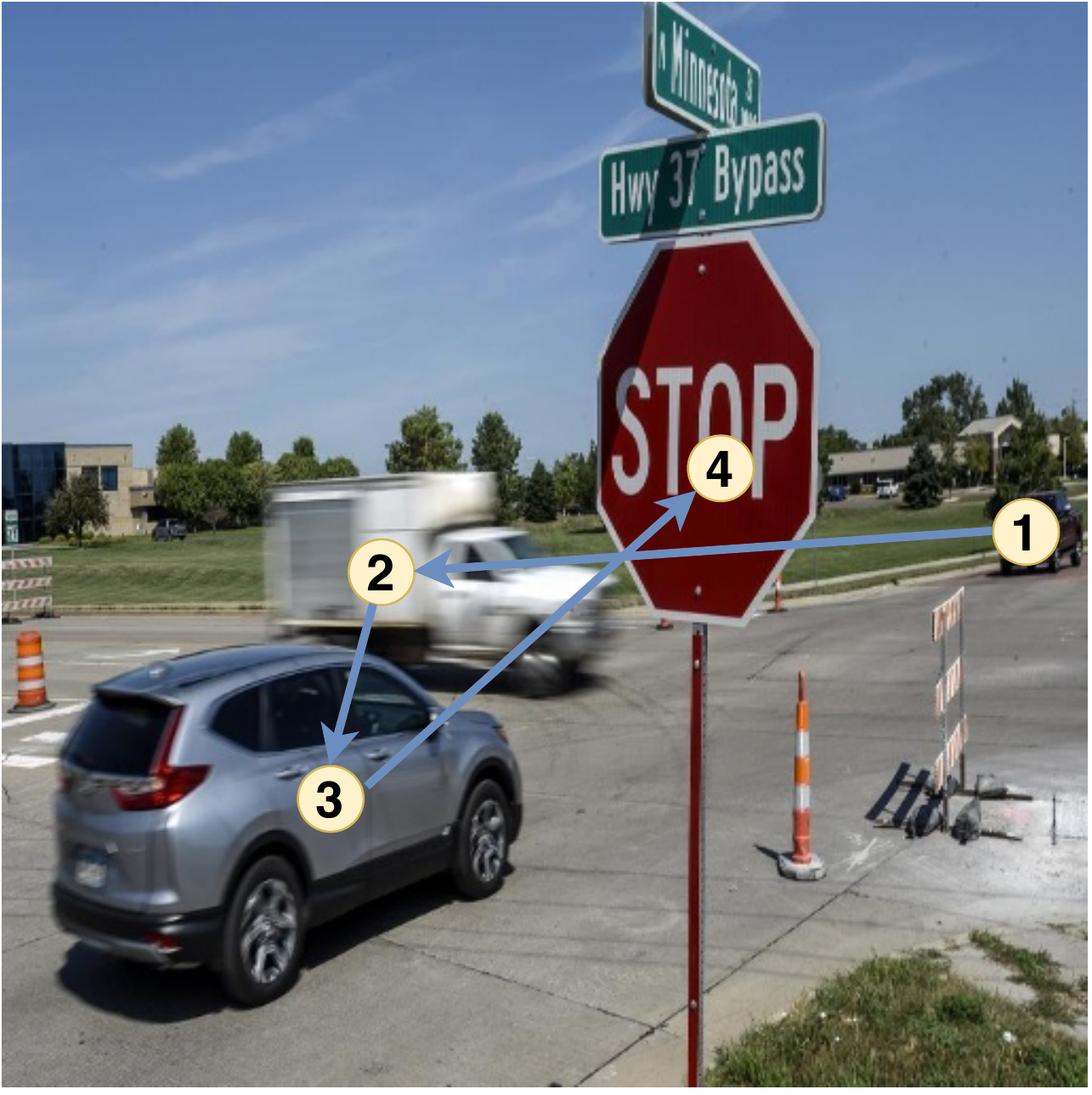}
        \caption{Stop}
    \end{subfigure}
    \begin{subfigure}{0.32\linewidth}
        \includegraphics[width=\linewidth]{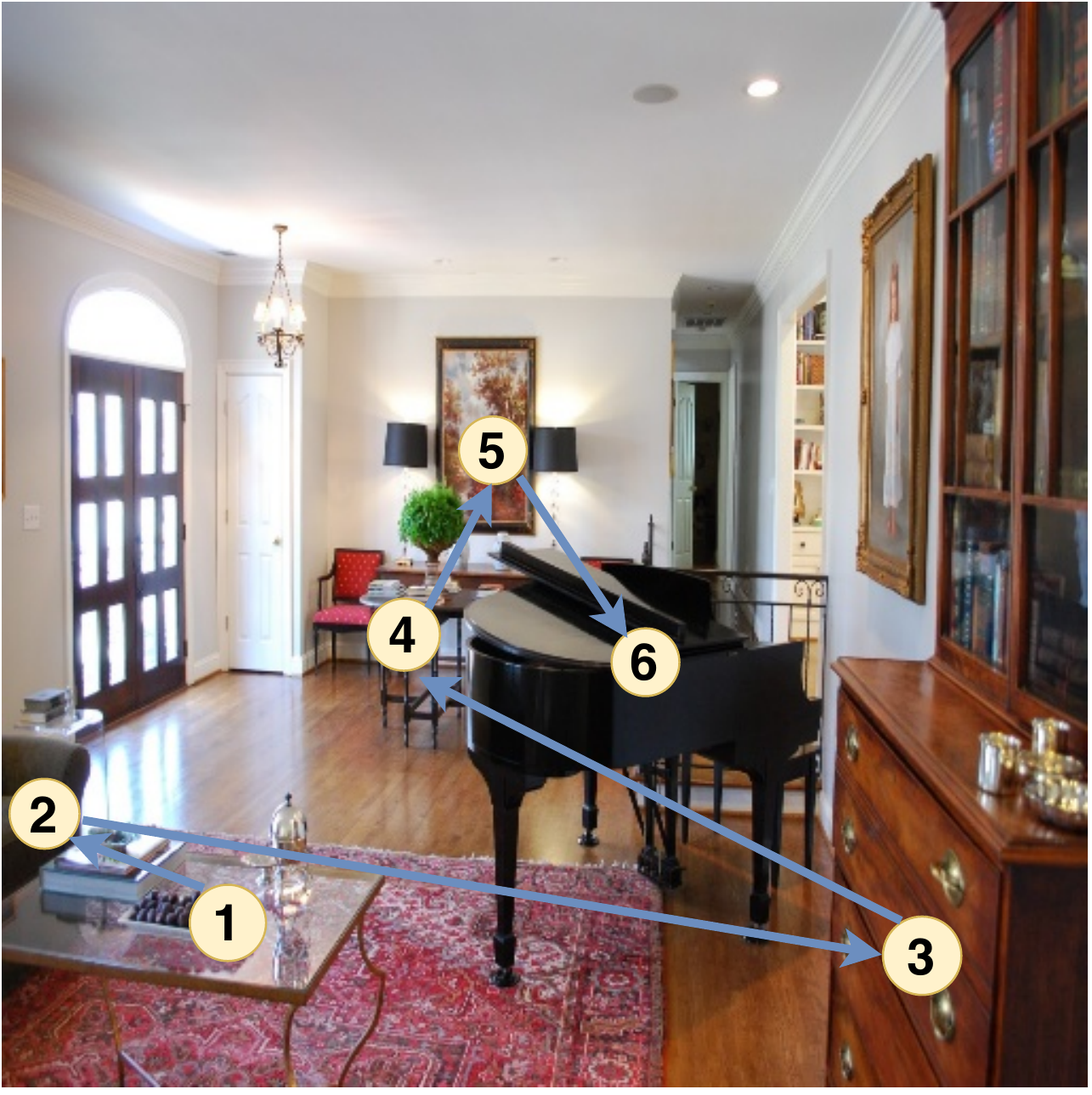}
        \caption{Piano}
    \end{subfigure}
    \caption{Reward maps learned by the our model for three different search targets which are piazza, stop signal and piano respectively in the context of Scanpath Prediction. The number means user's vision trajectory for searching target item which is the largest number refers to. For example, in (a) 2 represents piazza, in (b) 4 represents to stop sign and in (c), 6 represents to the piano. In addition, the hotmap represent the highest reward area which will be awarded to agent.}
    \label{fig:reward}
\end{figure}

Explainability plays a crucial role on the understanding of the decision-making process.
By visualizing the learned reward map, 
we show in this experiment our model can provide a certain level of interpretability.
We evaluate the explanability for our model in the scanpath prediction scenario.
Fig.~\ref{fig:reward} shows that the reward maps recovered by the our model depend heavily on the category of the search target. In the first image, the highest reward is assigned to the piazza when drinking beers. Similarly, the searching of road signal on the road, the stop signal get almost all of the reward while the car get only a few.

\subsection{Ablation Study}
\begin{table*}[!ht]
\caption{CTR for Different Parameter Settings for GAE and PPO with 95\% Confidence Interval}
\centering
\begin{tabular}{c|c|c|c|c|c|c|c}
    \hline
    \multicolumn{2}{c|}{\multirow{2}{*}{}} & \multicolumn{6}{c}{GAE: $\lambda_g$}      \\ \cline{3-8} 
    \multicolumn{2}{c|}{}                 & 0.94 & 0.95 & 0.96 & 0.97 & 0.98 & 0.99 \\ \hline
    \multirow{6}{*}{PPO: $\epsilon$} 
    & 0.05& 0.630 $\pm$ 0.063& 0.632 $\pm$ 0.064& 0.633 $\pm$ 0.062 & 0.630 $\pm$ 0.059 & 0.626 $\pm$ 0.060 & 0.629 $\pm$ 0.059 \\
                &  0.10& 0.632 $\pm$ 0.062 & 0.635 $\pm$ 0.060& 0.636 $\pm$ 0.061 & 0.636 $\pm$ 0.058 & 0.634 $\pm$ 0.061 & 0.633 $\pm$ 0.060 \\ 
                &  0.15& 0.633 $\pm$ 0.060 & 0.635 $\pm$ 0.061& 0.639 $\pm$ 0.061 & 0.640 $\pm$ 0.057 & 0.639 $\pm$ 0.059 & 0.638 $\pm$ 0.061 \\ 
                &  0.20& 0.634 $\pm$ 0.060 & 0.636 $\pm$ 0.060& 0.641 $\pm$ 0.063 & \textbf{0.643 $\pm$ 0.061} & 0.643 $\pm$ 0.063 & 0.641 $\pm$ 0.058 \\ 
                &  0.25& 0.631 $\pm$ 0.061 & 0.635 $\pm$ 0.059& 0.636 $\pm$ 0.060 & 0.637 $\pm$ 0.060 & 0.636 $\pm$ 0.061 & 0.634 $\pm$ 0.059 \\ 
                & 0.30 & 0.630 $\pm$ 0.059 & 0.631 $\pm$ 0.061 & 0.632 $\pm$ 0.060 & 0.630 $\pm$ 0.059 & 0.630 $\pm$ 0.058 & 0.629 $\pm$ 0.050\\ \hline
\end{tabular}
\label{tab:result}
\end{table*}
We test using three different optimization strategies (DDPG, Adaptive KL Penalty Coefficient and Twin Delayed DDPG) to update the policy parameter $\theta$. (TD3)~\cite{fujimoto2018addressing}.
The Adaptive KL Penalty Coefficient is defined as:
\begin{align}
    L(\theta) = \mathbb{E}_{t}\Big[ \frac{\pi_{\theta'}(a_t|s_t)}{\pi_{\theta}(a_t|s_t)}A_t - \beta \text{KL}[\pi_{\theta}(\cdot|s_t),\pi_{\theta'}(\cdot|s_t)]\Big]
\end{align}
where the $\beta$ will be adjust dynamically by the following, 
\begin{align}
    \begin{cases} 
      \beta \leftarrow \beta/2 & d < d_{target} * 1.5  \\
      \beta \leftarrow \beta*2 & d >= d_{target} * 1.5 \\
   \end{cases}\\
   \nonumber\text{where } d = \mathbb{E}_{t}[\text{KL}[\pi_{\theta_{old}}(\cdot|s_t),\pi_{\theta}(\cdot|s_t)]]
\end{align}
We empirically choose coefficient $1.5$ and $2$
and select total waiting time, CTR, and cumulative probability as the evaluation metrics to compare the three optimization strategies for traffic signal control, recommendation system, and scanpath prediction, respectively. The results (shown in Fig.~\ref{fig:compare}) show our optimization method achieve a similar result as TD3 on all the three tasks but is better than TD3 on the recommendation task.
Hence, we want to conduct a further step about the parameter selection about the PPO and GAE which can be found on Table~\ref{tab:result}.

\section{Related Work}
User behavior tendency modeling has been an active topic in research, and most previous efforts have been focusing on feature engineering rather than an end-to-end learning structure.
Kim et al.~\cite{kim2003learning} considers long-term interest as a reasonable representation of general interest and acknowledges its importance for personalization services. On this basis, Liu et al.~\cite{liu2007framework} propose a framework that considers both long-term and short-term interest for user behavior modeling. Rather than establishing static models, Chung et al.~\cite{chung2011incremental} models long-term and short-term user profile scores to model user behaviors incrementally.
Recently, Song et al.~\cite{song2016multi} propose to jointly model long-term and short-term user interest for recommendation based on deep learning methods. Pi et al.~\cite{pi2019practice} further propose a MIMN model for sequential user behavior modeling. Despite good performance on their respective tasks, all the above methods are task-specific and lack generalization ability.

Reinforcement learning is widely used for user behavior modeling in recommendation systems. Zheng et al.~\cite{zheng2018drn} adopt deep Q-learning to build up user profile during the interaction process in a music recommendation system. Zou et al.~\cite{zou2020pseudo} improve the Q-learning structure to stabilized the reward function and make the recommendation robust.
~\cite{chen2020knowledge,zhao2020leveraging,wang2020kerl} apply reinforcement learning for extracting user's interest from a knowledge graph. Liu et al.~\cite{liu2020end} embed user's information into a latent space and conduct recommendation via deep reinforcement learning. Different from those mentioned works, Pan et al.~\cite{pan2019policy} applies the policy gradient directly to optimize the recommendation policy. Chen et al.~\cite{chen2019generative} integrates the GAN into the reinforcement learning framework so that user's side information to enrich the latent space to improve the recommendation accuracy. Shang et al.~\cite{shang2019environment} considers the environment co-founder factors and propose a multi-agent based reinforcement learning method for recommendation.
All the above studies require defining accurate reward functions, which are hard to obtain in the real world.

Inverse reinforcement learning emerges where reward functions cannot be defined~\cite{ng2000algorithms}. Lee et al.~\cite{lee2010learning} firstly use the inverse reinforcement learning to learn user's behavior styles. However, general inverse reinforcement learning is computationally expensive. Ho et al.~\cite{ho2016generative} propose a generative reinforcement learning approach to improve efficiency. Fu et al.~\cite{fu2018learning} further extend the idea to a general form to obtain a more stable reward function. Kostrikov et al.~\cite{kostrikov2018discriminatoractorcritic} find a generative method may suffer instability in training, which can be relieved by using EM-distance instead of JS-divergence. Yang et al.~\cite{yang2020predicting} first introduces the inverse reinforcement learning into the scanpath prediction and demonstrate the superior performance. IRL demonstrates the huge potential and widely used in robot learning as it can empower agent to learn from the demonstration in different environments and tasks without dramatical exploration about the environment or familiar with the tasks. Chen et al.~\cite{chen2020generative} expands this idea into recommender system and shows the feasibility of IRL in recommendation task.  

\section{Conclusion and Future Work}
In this paper, we propose a new method based on \textit{advantage actor-critic network} with inverse reinforcement learning for user behavior modeling, to overcome the adverse impact caused by an inaccurate reward function.
In particular, we use the wasserstein GAN instead of GAN to increase training stability and a reply buffer for off-policy learning to increase sample efficiency.

Comparison of our method with several state-of-the-art methods in three different scenarios (namely traffic signal control, recommendation system, and scanpath prediction) demonstrate our method's feasibility in those scenarios and superior performance to baseline methods. 

Experience replay can boost the sample efficiency by switching the sampling process from the environment to replay buffer. However, it is not ideal as some tasks may introduce a giant state and action spaces such as recommendation. Sampling from such giant state and action spaces are not efficient. Moreover, not every experience are useful even it comes from the demonstration. The major reason is that expert demonstrations are sampled from replay buffer randomly and may orthogonal with the current state and lead to opposed actions. The possible solutions including, the state-aware experience replay method or prioritized  experience replay based methods~\cite{schaul2015prioritized}.
Another potential improvement leads by the Wasserstein GAN as the Lipschitz constraint is hard to enforce and may lead to model converge issue.
\bibliographystyle{IEEETran}
\bibliography{sample-base}

\begin{thebibliography}{10}
\providecommand{\url}[1]{#1}
\csname url@samestyle\endcsname
\providecommand{\newblock}{\relax}
\providecommand{\bibinfo}[2]{#2}
\providecommand{\BIBentrySTDinterwordspacing}{\spaceskip=0pt\relax}
\providecommand{\BIBentryALTinterwordstretchfactor}{4}
\providecommand{\BIBentryALTinterwordspacing}{\spaceskip=\fontdimen2\font plus
\BIBentryALTinterwordstretchfactor\fontdimen3\font minus
  \fontdimen4\font\relax}
\providecommand{\BIBforeignlanguage}[2]{{%
\expandafter\ifx\csname l@#1\endcsname\relax
\typeout{** WARNING: IEEEtran.bst: No hyphenation pattern has been}%
\typeout{** loaded for the language `#1'. Using the pattern for}%
\typeout{** the default language instead.}%
\else
\language=\csname l@#1\endcsname
\fi
#2}}
\providecommand{\BIBdecl}{\relax}
\BIBdecl

\bibitem{hong2009context}
J.~Hong, E.-H. Suh, J.~Kim, and S.~Kim, ``Context-aware system for proactive
  personalized service based on context history,'' \emph{Expert Systems with
  Applications}, vol.~36, no.~4, pp. 7448--7457, 2009.

\bibitem{zheng2018drn}
G.~Zheng, F.~Zhang, Z.~Zheng, Y.~Xiang, N.~J. Yuan, X.~Xie, and Z.~Li, ``Drn: A
  deep reinforcement learning framework for news recommendation,'' in
  \emph{Proceedings of the 2018 World Wide Web Conference}, 2018, pp. 167--176.

\bibitem{zhang2019deep}
S.~Zhang, L.~Yao, A.~Sun, and Y.~Tay, ``Deep learning based recommender system:
  A survey and new perspectives,'' \emph{ACM Computing Surveys (CSUR)},
  vol.~52, no.~1, pp. 1--38, 2019.

\bibitem{hu2017diversifying}
L.~Hu, L.~Cao, S.~Wang, G.~Xu, J.~Cao, and Z.~Gu, ``Diversifying personalized
  recommendation with user-session context.'' in \emph{IJCAI}, 2017, pp.
  1858--1864.

\bibitem{xu2020contextual}
X.~Xu, F.~Dong, Y.~Li, S.~He, and X.~Li, ``Contextual-bandit based personalized
  recommendation with time-varying user interests.'' in \emph{AAAI}, 2020, pp.
  6518--6525.

\bibitem{wang2014exploration}
X.~Wang, Y.~Wang, D.~Hsu, and Y.~Wang, ``Exploration in interactive
  personalized music recommendation: a reinforcement learning approach,''
  \emph{ACM Transactions on Multimedia Computing, Communications, and
  Applications (TOMM)}, vol.~11, no.~1, pp. 1--22, 2014.

\bibitem{kim2003learning}
H.~R. Kim and P.~K. Chan, ``Learning implicit user interest hierarchy for
  context in personalization,'' in \emph{Proceedings of the 8th international
  conference on Intelligent user interfaces}, 2003, pp. 101--108.

\bibitem{sheridan2016human}
T.~B. Sheridan, ``Human--robot interaction: status and challenges,''
  \emph{Human factors}, vol.~58, no.~4, pp. 525--532, 2016.

\bibitem{schmerling2018multimodal}
E.~Schmerling, K.~Leung, W.~Vollprecht, and M.~Pavone, ``Multimodal
  probabilistic model-based planning for human-robot interaction,'' in
  \emph{2018 IEEE International Conference on Robotics and Automation
  (ICRA)}.\hskip 1em plus 0.5em minus 0.4em\relax IEEE, 2018, pp. 1--9.

\bibitem{siam2019video}
M.~Siam, C.~Jiang, S.~Lu, L.~Petrich, M.~Gamal, M.~Elhoseiny, and M.~Jagersand,
  ``Video object segmentation using teacher-student adaptation in a human robot
  interaction (hri) setting,'' in \emph{2019 International Conference on
  Robotics and Automation (ICRA)}.\hskip 1em plus 0.5em minus 0.4em\relax IEEE,
  2019, pp. 50--56.

\bibitem{badii2017user}
C.~Badii, P.~Bellini, D.~Cenni, A.~Difino, M.~Paolucci, and P.~Nesi, ``User
  engagement engine for smart city strategies,'' in \emph{2017 IEEE
  International Conference on Smart Computing (SMARTCOMP)}.\hskip 1em plus
  0.5em minus 0.4em\relax IEEE, 2017, pp. 1--7.

\bibitem{bellini2017wi}
P.~Bellini, D.~Cenni, P.~Nesi, and I.~Paoli, ``Wi-fi based city users’
  behaviour analysis for smart city,'' \emph{Journal of Visual Languages \&
  Computing}, vol.~42, pp. 31--45, 2017.

\bibitem{seko2011group}
S.~Seko, T.~Yagi, M.~Motegi, and S.~Muto, ``Group recommendation using feature
  space representing behavioral tendency and power balance among members,'' in
  \emph{Proceedings of the fifth ACM conference on Recommender systems}, 2011,
  pp. 101--108.

\bibitem{shi2014collaborative}
Y.~Shi, M.~Larson, and A.~Hanjalic, ``Collaborative filtering beyond the
  user-item matrix: A survey of the state of the art and future challenges,''
  \emph{ACM Computing Surveys (CSUR)}, vol.~47, no.~1, pp. 1--45, 2014.

\bibitem{yang2020predicting}
Z.~Yang, L.~Huang, Y.~Chen, Z.~Wei, S.~Ahn, G.~Zelinsky, D.~Samaras, and
  M.~Hoai, ``Predicting goal-directed human attention using inverse
  reinforcement learning,'' in \emph{Proceedings of the IEEE/CVF Conference on
  Computer Vision and Pattern Recognition}, 2020, pp. 193--202.

\bibitem{bazzan2009opportunities}
A.~L. Bazzan, ``Opportunities for multiagent systems and multiagent
  reinforcement learning in traffic control,'' \emph{Autonomous Agents and
  Multi-Agent Systems}, vol.~18, no.~3, p. 342, 2009.

\bibitem{liu2018interactive}
H.~Liu, Y.~Zhang, W.~Si, X.~Xie, Y.~Zhu, and S.-C. Zhu, ``Interactive robot
  knowledge patching using augmented reality,'' in \emph{2018 IEEE
  International Conference on Robotics and Automation (ICRA)}.\hskip 1em plus
  0.5em minus 0.4em\relax IEEE, 2018, pp. 1947--1954.

\bibitem{chen2018stabilizing}
S.-Y. Chen, Y.~Yu, Q.~Da, J.~Tan, H.-K. Huang, and H.-H. Tang, ``Stabilizing
  reinforcement learning in dynamic environment with application to online
  recommendation,'' in \emph{Proceedings of the 24th ACM SIGKDD International
  Conference on Knowledge Discovery \& Data Mining}, 2018, pp. 1187--1196.

\bibitem{chen2019large}
H.~Chen, X.~Dai, H.~Cai, W.~Zhang, X.~Wang, R.~Tang, Y.~Zhang, and Y.~Yu,
  ``Large-scale interactive recommendation with tree-structured policy
  gradient,'' in \emph{Proceedings of the AAAI Conference on Artificial
  Intelligence}, vol.~33, 2019, pp. 3312--3320.

\bibitem{ng2000algorithms}
A.~Y. Ng, S.~J. Russell \emph{et~al.}, ``Algorithms for inverse reinforcement
  learning.'' in \emph{Icml}, vol.~1, 2000, p.~2.

\bibitem{ho2016generative}
J.~Ho and S.~Ermon, ``Generative adversarial imitation learning,'' in
  \emph{Advances in neural information processing systems}, 2016, pp.
  4565--4573.

\bibitem{chen2020generative}
X.~Chen, L.~Yao, A.~Sun, X.~Wang, X.~Xu, and L.~Zhu, ``Generative inverse deep
  reinforcement learning for online recommendation,'' \emph{arXiv preprint
  arXiv:2011.02248}, 2020.

\bibitem{abbeel2004apprenticeship}
P.~Abbeel and A.~Y. Ng, ``Apprenticeship learning via inverse reinforcement
  learning,'' in \emph{Proceedings of the twenty-first international conference
  on Machine learning}.\hskip 1em plus 0.5em minus 0.4em\relax ACM, 2004, p.~1.

\bibitem{syed2008apprenticeship}
U.~Syed, M.~Bowling, and R.~E. Schapire, ``Apprenticeship learning using linear
  programming,'' in \emph{Proceedings of the 25th international conference on
  Machine learning}, 2008, pp. 1032--1039.

\bibitem{syed2008game}
U.~Syed and R.~E. Schapire, ``A game-theoretic approach to apprenticeship
  learning,'' in \emph{Advances in neural information processing systems},
  2008, pp. 1449--1456.

\bibitem{bloem2014infinite}
M.~Bloem and N.~Bambos, ``Infinite time horizon maximum causal entropy inverse
  reinforcement learning,'' in \emph{53rd IEEE Conference on Decision and
  Control}.\hskip 1em plus 0.5em minus 0.4em\relax IEEE, 2014, pp. 4911--4916.

\bibitem{ziebart2010modeling}
B.~D. Ziebart, J.~A. Bagnell, and A.~K. Dey, ``Modeling interaction via the
  principle of maximum causal entropy,'' 2010.

\bibitem{goodfellow2014generative}
I.~Goodfellow, J.~Pouget-Abadie, M.~Mirza, B.~Xu, D.~Warde-Farley, S.~Ozair,
  A.~Courville, and Y.~Bengio, ``Generative adversarial nets,'' in
  \emph{Advances in neural information processing systems}, 2014, pp.
  2672--2680.

\bibitem{nguyen2009surrogate}
X.~Nguyen, M.~J. Wainwright, M.~I. Jordan \emph{et~al.}, ``On surrogate loss
  functions and f-divergences,'' \emph{The Annals of Statistics}, vol.~37,
  no.~2, pp. 876--904, 2009.

\bibitem{fu2018learning}
\BIBentryALTinterwordspacing
J.~Fu, K.~Luo, and S.~Levine, ``Learning robust rewards with adverserial
  inverse reinforcement learning,'' in \emph{International Conference on
  Learning Representations}, 2018. [Online]. Available:
  \url{https://openreview.net/forum?id=rkHywl-A-}
\BIBentrySTDinterwordspacing

\bibitem{kostrikov2018discriminatoractorcritic}
\BIBentryALTinterwordspacing
I.~Kostrikov, K.~K. Agrawal, D.~Dwibedi, S.~Levine, and J.~Tompson,
  ``Discriminator-actor-critic: Addressing sample inefficiency and reward bias
  in adversarial imitation learning,'' in \emph{International Conference on
  Learning Representations}, 2019. [Online]. Available:
  \url{https://openreview.net/forum?id=Hk4fpoA5Km}
\BIBentrySTDinterwordspacing

\bibitem{mnih2016asynchronous}
V.~Mnih, A.~P. Badia, M.~Mirza, A.~Graves, T.~Lillicrap, T.~Harley, D.~Silver,
  and K.~Kavukcuoglu, ``Asynchronous methods for deep reinforcement learning,''
  in \emph{International conference on machine learning}, 2016, pp. 1928--1937.

\bibitem{konda2000actor}
V.~R. Konda and J.~N. Tsitsiklis, ``Actor-critic algorithms,'' in
  \emph{Advances in neural information processing systems}, 2000, pp.
  1008--1014.

\bibitem{chen2020knowledge}
X.~Chen, C.~Huang, L.~Yao, X.~Wang, W.~Liu, and W.~Zhang, ``Knowledge-guided
  deep reinforcement learning for interactive recommendation,'' \emph{arXiv
  preprint arXiv:2004.08068}, 2020.

\bibitem{liu2020end}
F.~Liu, H.~Guo, X.~Li, R.~Tang, Y.~Ye, and X.~He, ``End-to-end deep
  reinforcement learning based recommendation with supervised embedding,'' in
  \emph{Proceedings of the 13th International Conference on Web Search and Data
  Mining}, 2020, pp. 384--392.

\bibitem{kingma2014adam}
D.~P. Kingma and J.~Ba, ``Adam: A method for stochastic optimization,''
  \emph{arXiv preprint arXiv:1412.6980}, 2014.

\bibitem{wang2016sample}
Z.~Wang, V.~Bapst, N.~Heess, V.~Mnih, R.~Munos, K.~Kavukcuoglu, and
  N.~de~Freitas, ``Sample efficient actor-critic with experience replay,''
  \emph{arXiv preprint arXiv:1611.01224}, 2016.

\bibitem{arjovsky2017wasserstein}
M.~Arjovsky, S.~Chintala, and L.~Bottou, ``Wasserstein gan,'' \emph{arXiv
  preprint arXiv:1701.07875}, 2017.

\bibitem{gulrajani2017improved}
I.~Gulrajani, F.~Ahmed, M.~Arjovsky, V.~Dumoulin, and A.~C. Courville,
  ``Improved training of wasserstein gans,'' in \emph{Advances in neural
  information processing systems}, 2017, pp. 5767--5777.

\bibitem{villani2008optimal}
C.~Villani, \emph{Optimal transport: old and new}.\hskip 1em plus 0.5em minus
  0.4em\relax Springer Science \& Business Media, 2008, vol. 338.

\bibitem{lucic2018gans}
M.~Lucic, K.~Kurach, M.~Michalski, S.~Gelly, and O.~Bousquet, ``Are gans
  created equal? a large-scale study,'' in \emph{Advances in neural information
  processing systems}, 2018, pp. 700--709.

\bibitem{schulman2015trust}
J.~Schulman, S.~Levine, P.~Abbeel, M.~Jordan, and P.~Moritz, ``Trust region
  policy optimization,'' in \emph{International conference on machine
  learning}, 2015, pp. 1889--1897.

\bibitem{schulman2015high}
J.~Schulman, P.~Moritz, S.~Levine, M.~Jordan, and P.~Abbeel, ``High-dimensional
  continuous control using generalized advantage estimation,'' \emph{arXiv
  preprint arXiv:1506.02438}, 2015.

\bibitem{schulman2017proximal}
J.~Schulman, F.~Wolski, P.~Dhariwal, A.~Radford, and O.~Klimov, ``Proximal
  policy optimization algorithms,'' \emph{arXiv preprint arXiv:1707.06347},
  2017.

\bibitem{paszke2019pytorch}
A.~Paszke, S.~Gross, F.~Massa, A.~Lerer, J.~Bradbury, G.~Chanan, T.~Killeen,
  Z.~Lin, N.~Gimelshein, L.~Antiga \emph{et~al.}, ``Pytorch: An imperative
  style, high-performance deep learning library,'' in \emph{Advances in neural
  information processing systems}, 2019, pp. 8026--8037.

\bibitem{SUMO2018}
\BIBentryALTinterwordspacing
P.~A. Lopez, M.~Behrisch, L.~Bieker-Walz, J.~Erdmann, Y.-P. Fl{\"o}tter{\"o}d,
  R.~Hilbrich, L.~L{\"u}cken, J.~Rummel, P.~Wagner, and E.~Wie{\ss}ner,
  ``Microscopic traffic simulation using sumo,'' in \emph{The 21st IEEE
  International Conference on Intelligent Transportation Systems}.\hskip 1em
  plus 0.5em minus 0.4em\relax IEEE, 2018. [Online]. Available:
  \url{https://elib.dlr.de/124092/}
\BIBentrySTDinterwordspacing

\bibitem{gao2018reinforcement}
\BIBentryALTinterwordspacing
Y.~Gao, H.~Xu, J.~Lin, F.~Yu, S.~Levine, and T.~Darrell, ``Reinforcement
  learning from imperfect demonstrations,'' 2018. [Online]. Available:
  \url{https://openreview.net/forum?id=BJJ9bz-0-}
\BIBentrySTDinterwordspacing

\bibitem{lillicrap2015continuous}
T.~P. Lillicrap, J.~J. Hunt, A.~Pritzel, N.~Heess, T.~Erez, Y.~Tassa,
  D.~Silver, and D.~Wierstra, ``Continuous control with deep reinforcement
  learning,'' \emph{arXiv preprint arXiv:1509.02971}, 2015.

\bibitem{rummery1994line}
G.~A. Rummery and M.~Niranjan, \emph{On-line Q-learning using connectionist
  systems}.\hskip 1em plus 0.5em minus 0.4em\relax University of Cambridge,
  Department of Engineering Cambridge, UK, 1994, vol.~37.

\bibitem{shi2019virtual}
J.-C. Shi, Y.~Yu, Q.~Da, S.-Y. Chen, and A.-X. Zeng, ``Virtual-taobao:
  Virtualizing real-world online retail environment for reinforcement
  learning,'' in \emph{Proceedings of the AAAI Conference on Artificial
  Intelligence}, vol.~33, 2019, pp. 4902--4909.

\bibitem{bai2019model}
X.~Bai, J.~Guan, and H.~Wang, ``A model-based reinforcement learning with
  adversarial training for online recommendation,'' in \emph{Advances in Neural
  Information Processing Systems}, 2019, pp. 10\,735--10\,746.

\bibitem{pan2019policy}
F.~Pan, Q.~Cai, P.~Tang, F.~Zhuang, and Q.~He, ``Policy gradients for
  contextual recommendations,'' in \emph{The World Wide Web Conference}, 2019,
  pp. 1421--1431.

\bibitem{chen2019generative}
X.~Chen, S.~Li, H.~Li, S.~Jiang, Y.~Qi, and L.~Song, ``Generative adversarial
  user model for reinforcement learning based recommendation system,'' in
  \emph{International Conference on Machine Learning}, 2019, pp. 1052--1061.

\bibitem{chen2015deepdriving}
C.~Chen, A.~Seff, A.~Kornhauser, and J.~Xiao, ``Deepdriving: Learning
  affordance for direct perception in autonomous driving,'' in
  \emph{Proceedings of the IEEE International Conference on Computer Vision},
  2015, pp. 2722--2730.

\bibitem{ballas2015delving}
N.~Ballas, L.~Yao, C.~Pal, and A.~Courville, ``Delving deeper into
  convolutional networks for learning video representations,'' \emph{arXiv
  preprint arXiv:1511.06432}, 2015.

\bibitem{fujimoto2018addressing}
S.~Fujimoto, H.~Van~Hoof, and D.~Meger, ``Addressing function approximation
  error in actor-critic methods,'' \emph{arXiv preprint arXiv:1802.09477},
  2018.

\bibitem{liu2007framework}
H.~Liu and M.~Zamanian, ``Framework for selecting and delivering advertisements
  over a network based on combined short-term and long-term user behavioral
  interests,'' Mar.~15 2007, uS Patent App. 11/225,238.

\bibitem{chung2011incremental}
C.~Y. Chung, A.~Gupta, J.~M. Koran, L.-J. Lin, and H.~Yin, ``Incremental update
  of long-term and short-term user profile scores in a behavioral targeting
  system,'' Mar.~8 2011, uS Patent 7,904,448.

\bibitem{song2016multi}
Y.~Song, A.~M. Elkahky, and X.~He, ``Multi-rate deep learning for temporal
  recommendation,'' in \emph{Proceedings of the 39th International ACM SIGIR
  conference on Research and Development in Information Retrieval}, 2016, pp.
  909--912.

\bibitem{pi2019practice}
Q.~Pi, W.~Bian, G.~Zhou, X.~Zhu, and K.~Gai, ``Practice on long sequential user
  behavior modeling for click-through rate prediction,'' in \emph{Proceedings
  of the 25th ACM SIGKDD International Conference on Knowledge Discovery \&
  Data Mining}, 2019, pp. 2671--2679.

\bibitem{zou2020pseudo}
L.~Zou, L.~Xia, P.~Du, Z.~Zhang, T.~Bai, W.~Liu, J.-Y. Nie, and D.~Yin,
  ``Pseudo dyna-q: A reinforcement learning framework for interactive
  recommendation,'' in \emph{Proceedings of the 13th International Conference
  on Web Search and Data Mining}, 2020, pp. 816--824.

\bibitem{zhao2020leveraging}
K.~Zhao, X.~Wang, Y.~Zhang, L.~Zhao, Z.~Liu, C.~Xing, and X.~Xie, ``Leveraging
  demonstrations for reinforcement recommendation reasoning over knowledge
  graphs,'' in \emph{Proceedings of the 43rd International ACM SIGIR Conference
  on Research and Development in Information Retrieval}, 2020, pp. 239--248.

\bibitem{wang2020kerl}
P.~Wang, Y.~Fan, L.~Xia, W.~X. Zhao, S.~Niu, and J.~Huang, ``Kerl: A
  knowledge-guided reinforcement learning model for sequential
  recommendation,'' in \emph{Proceedings of the 43rd International ACM SIGIR
  Conference on Research and Development in Information Retrieval}, 2020, pp.
  209--218.

\bibitem{shang2019environment}
W.~Shang, Y.~Yu, Q.~Li, Z.~Qin, Y.~Meng, and J.~Ye, ``Environment
  reconstruction with hidden confounders for reinforcement learning based
  recommendation,'' in \emph{Proceedings of the 25th ACM SIGKDD International
  Conference on Knowledge Discovery \& Data Mining}, 2019, pp. 566--576.

\bibitem{lee2010learning}
S.~J. Lee and Z.~Popovi{\'c}, ``Learning behavior styles with inverse
  reinforcement learning,'' \emph{ACM transactions on graphics (TOG)}, vol.~29,
  no.~4, pp. 1--7, 2010.

\bibitem{schaul2015prioritized}
T.~Schaul, J.~Quan, I.~Antonoglou, and D.~Silver, ``Prioritized experience
  replay,'' \emph{arXiv preprint arXiv:1511.05952}, 2015.

\end{thebibliography}
%




\end{document}